    \documentclass[letterpaper]{article} 
    \usepackage[]{aaai2026}  
    \usepackage{times}  
    \usepackage{helvet}  
    \usepackage{courier}  
    \usepackage[hyphens]{url}  
    \usepackage{graphicx} 
    \usepackage{amssymb}
    \usepackage{natbib}  
    \usepackage{caption} 
    \usepackage{algorithm}
    \usepackage{algorithmic}
    \usepackage{booktabs}
    \usepackage{subfigure}
    \usepackage{newfloat}
    \usepackage{listings}
    \DeclareCaptionStyle{ruled}{labelfont=normalfont,labelsep=colon,strut=off} 
    \floatstyle{ruled}
    \newfloat{listing}{tb}{lst}{}
    \floatname{listing}{Listing}
    \nocopyright 
    
    \title{Finetuning-Free Personalization of Text to Image Generation via Hypernetworks}
    \author {
        Sagar Shrestha\textsuperscript{\rm 1,2}\footnote{Work done during an internship at Samsung AI Center Mountain View.},
        Gopal Sharma\textsuperscript{\rm 1},
        Luowei Zhou\textsuperscript{\rm 1},
        Suren Kumar\textsuperscript{\rm 1}
    }
    \affiliations {
        \textsuperscript{\rm 1}AI Center-Mountain View, Samsung Electronics\\
        \textsuperscript{\rm 2}Oregon State University\\
        shressag@oregonstate.edu, \{gopal.sharma,luowei.zhou,suren.kumar\}@samsung.com
    }

    \usepackage{bibentry}
    
    \usepackage{steinmetz}

\newcommand{\h}{\boldsymbol{h}}

\newcommand{\s}{\boldsymbol{s}}

\newcommand{\x}{\boldsymbol{x}}

\newcommand{\cL}{\mathcal{L}}

\usepackage{extarrows}

\DeclareMathOperator*{\minimize}{\textrm{min}}

\usepackage{xcolor}
\usepackage{psfrag,framed}
\usepackage{lipsum}
\usepackage{chapterbib}
\PassOptionsToPackage{normalem}{ulem}
\usepackage{ulem}
\definecolor{shadecolor}{RGB}{220,220,220}

\newcommand{\bm}{\boldsymbol}

\begin{document}
    
    \maketitle
    
    \begin{abstract}

    Personalizing text-to-image diffusion models has traditionally relied on subject-specific fine-tuning approaches such as DreamBooth~\cite{ruiz2023dreambooth}, which are computationally expensive and slow at inference. Recent adapter- and encoder-based methods attempt to reduce this overhead but still depend on additional fine-tuning or large backbone models for satisfactory results. In this work, we revisit an orthogonal direction: \emph{fine-tuning-free} personalization via Hypernetworks that predict LoRA-adapted weights directly from subject images. Prior hypernetwork-based approaches, however, suffer from costly data generation or unstable attempts to mimic base model optimization trajectories. We address these limitations with an \emph{end-to-end} training objective, stabilized by a simple output regularization, yielding reliable and effective hypernetworks. Our method removes the need for per-subject optimization at test time while preserving both subject fidelity and prompt alignment. To further enhance compositional generalization at inference time, we introduce Hybrid-Model Classifier-Free Guidance (HM-CFG), which combines the compositional strengths of the base diffusion model with the subject fidelity of personalized models during sampling. Extensive experiments on CelebA-HQ, AFHQ-v2, and DreamBench demonstrate that our approach achieves strong personalization performance and highlights the promise of hypernetworks as a scalable and effective direction for open-category personalization.
    \end{abstract}
    
    \section{Introduction}
    In this work, we are interested in personalized generation--producing images of a specific subject instance such as a pet, a face, or a user-defined object conditioned on a prompt. 
    The rapid progress of diffusion-based text-to-image (T2I) models \cite{rombach2022high} have revolutionized generative image synthesis, enabling the creation of highly diverse and semantically coherent images from natural language prompts. 
    However, existing T2I models fall short at the task of personalization and requires additional adaptation.
    
    To address this limitation, early works on personalization, such as DreamBooth \cite{ruiz2023dreambooth} and custom diffusion \cite{kumari2023multi}, proposed fine-tuning a pre-trained diffusion model using a small set of subject images. These methods have proven effective in maintaining subject fidelity while preserving prompt compliance, making fine-tuning the promising approach for high-quality subject-driven generation. However, this comes at the cost of significant computational overhead. Each new subject requires several minutes of fine-tuning time on high-memory GPUs (e.g., 26GB for SDXL using fp32 precision),
    which severely limits their applicability in real-time or large-scale settings.

    To reduce this overhead, recent methods explore fine-tuning-free alternatives, including adapters \cite{ye2023ip}, encoders \cite{gal2023designing}, and prompt-based techniques \cite{kang2025flux}. These approaches condition the diffusion process using auxiliary embeddings or token substitutions, avoiding subject-specific optimization at inference. Yet, they often struggle with subject fidelity in challenging cases and rely either on light fine-tuning ($\sim$100 steps) to achieve acceptable alignment (see Table \ref{tab:ft_existing_table}), which undermines the goal of truly fine-tuning-free personalization, or inherent capabilities of very large base models such as FLUX \cite{shin2025large, kang2025flux}.
    
\begin{figure*}[t]
        \centering
        \subfigure[Hypernet Training]{
            \includegraphics[width=0.44\linewidth]{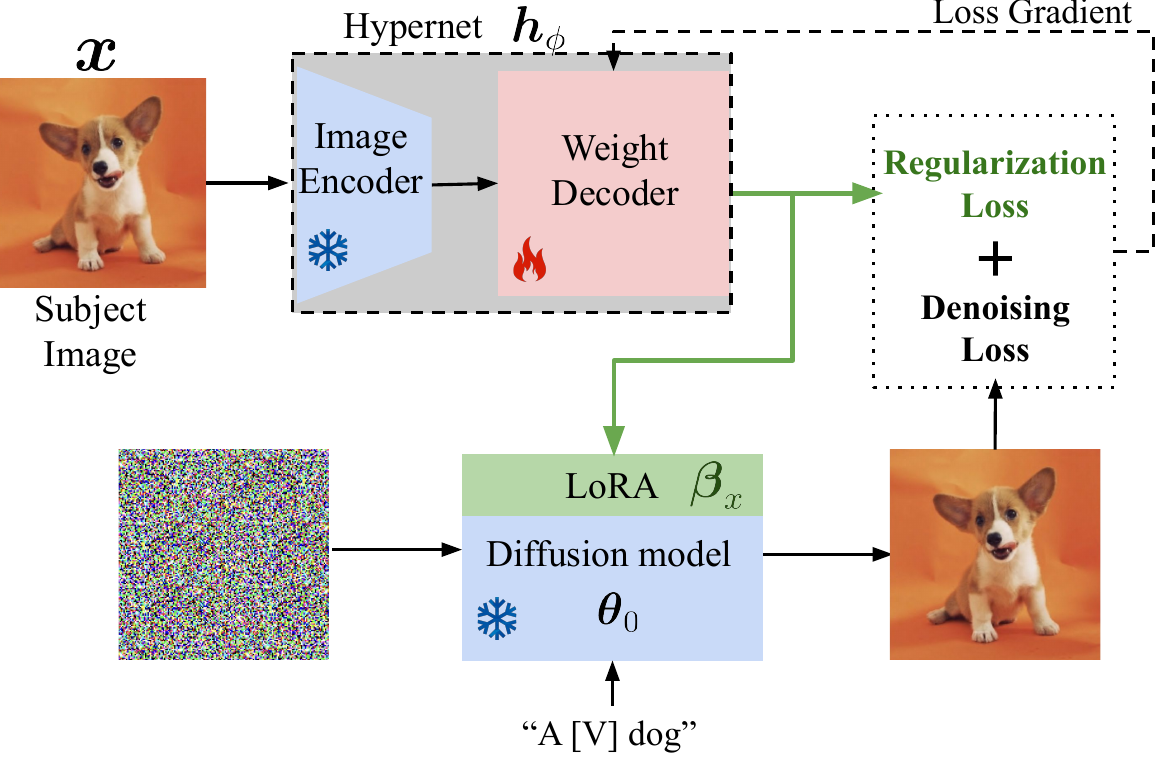}
            \label{fig:hypernet_illus}
        }
        \subfigure[HM-CFG Inference]{
            \includegraphics[width=0.53\linewidth]{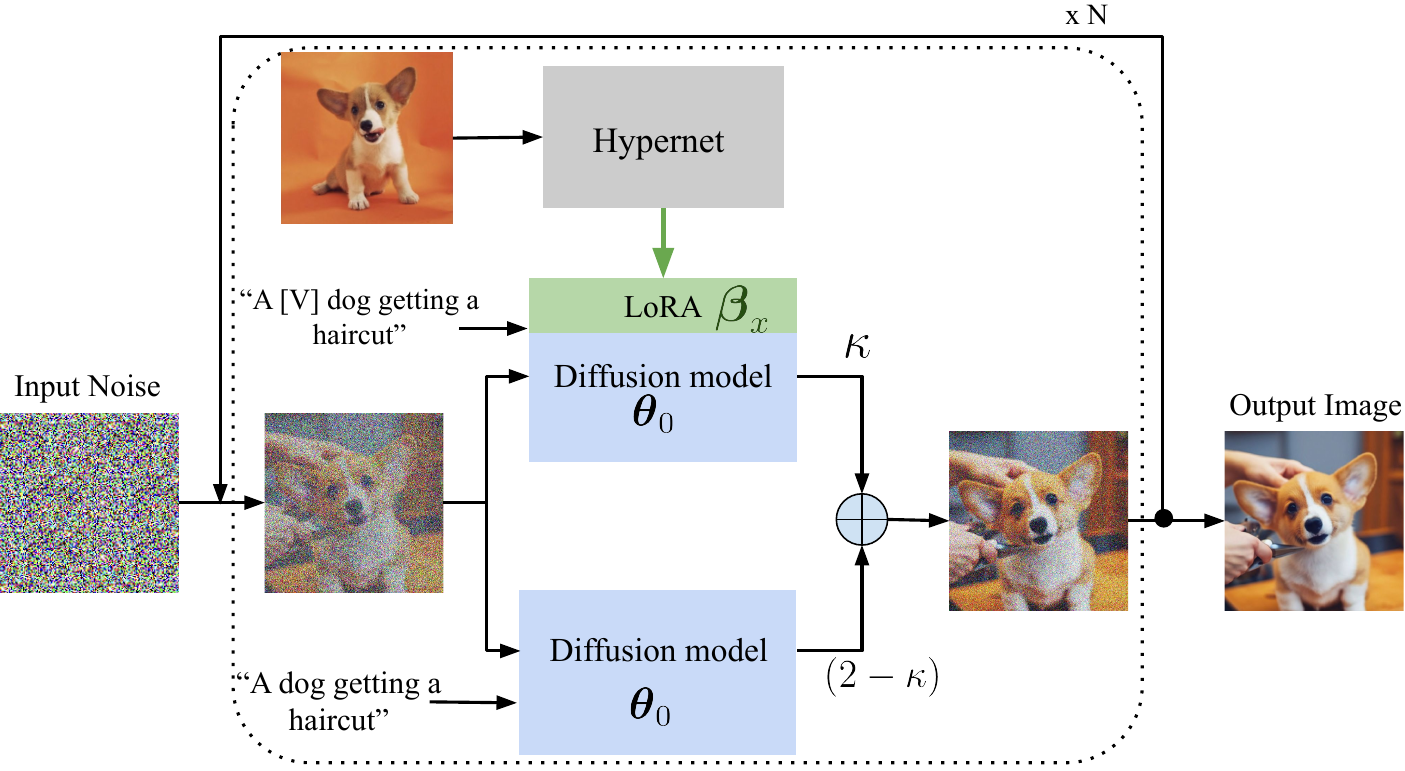}
        }
        \vspace{-3mm}
        \caption{\textbf{Overview of our approach.}
    \textbf{a)} Our proposed training pipeline for hypernetwork based personalization. A frozen image encoder processes the input image, and a trainable weight decoder predicts the corresponding LoRA parameters. These parameters are then used to adapt a frozen, pre-trained text-to-image diffusion model. The hypernetwork is optimized using a composite loss function that includes both a denoising diffusion term and a regularization term on the hypernetwork's output as shown in Sec~\ref{sec:end-to-end}. \textbf{b)} Our proposed inference approach using hybrid model based classifier-free guidance that combines base model and LoRA adapted model to improve compositional prompt adherence, as described in Sec~\ref{sec:hm-cfg}.
        }
        \vspace{-3mm}
    \end{figure*}

    An underexplored complementary direction lies in hypernetworks--auxiliary networks that generate parameters for a target model conditioned on input images. In T2I personalization, hypernetworks can be trained to predict fine-tuned weights (e.g., LoRA adapters) for a diffusion model directly from subject images, thereby amortizing the cost of per-subject optimization (see Fig. \ref{fig:hypernet_illus}). Prior work such as HyperDreamBooth~\cite{ruiz2023hyperdreambooth} has demonstrated the feasibility of this idea, but often depends on large collections of fine-tuned model weights and introduces inference-time adaptation costs~\cite{hedlin2025hypernet,ruiz2023hyperdreambooth}. Moreover, existing approaches have been largely restricted to “closed-category” settings such as human faces or pets.
    
    We propose a fully fine-tuning-free personalization framework based on \emph{end-to-end} hypernetwork training. Our hypernetwork is trained directly to predict LoRA parameters for a frozen base diffusion model using subject images as input. Unlike prior methods that rely on precomputed (image, fine-tuned-weights) pairs or mimic optimization trajectories, our approach avoids noisy supervision and decoupled training. A simple output regularization suffices to stabilize learning, enabling reliable and effective hypernetworks without requiring test-time optimization.

    Another fundamental issues associated with T2I personalization is that it tends to overfit to the small amount of provided subject images. This is characterized by poor prompt following which suggests that the models tend to forget general linguistic abilities acquired during T2I pre-training. Hence, as an additional contribution, we introduce Hybrid Model Classifier-Free Guidance (HM-CFG), a modification of the popular CFG based sampling designed for personalized diffusion models. It combines the prompt following ability of the base diffusion model with the subject generation ability of fine-tuned diffusion models via an efficient guidance scheme at inference time. In principle, this method can be applied for inference with any personalized diffusion model. We demonstrate its effectiveness in providing user control between subject and prompt fidelity for various datasets.

    In summary, our contributions are as follows:
    \begin{enumerate}
        \item We propose an end-to-end training approach for hypernetwork that predicts subject-specific LoRA weights for text-to-image diffusion models, eliminating the need for test-time fine-tuning.
    
        \item We introduce a regularization strategy to stabilize end-to-end training and prevent overfitting of hypernetwork.
    
        \item We design an inference strategy called hybrid model classifier-free guidance (HM-CFG) mechanism that improves compositional prompt adherence while preserving subject fidelity.
    
        \item We conduct extensive experiments across benchmark datasets and show state-of-the-art performance among existing personalization methods using stable diffusion models in both open and closed category settings.
    \end{enumerate}
    
    \section{Related Works}
    
    \subsection{Text-to-Image Synthesis}
    The landscape of generative AI has been reshaped by text-to-image (T2I) diffusion models. Seminal works like DALL-E 2 \cite{ramesh2022hierarchical}, Imagen \cite{saharia2022photorealistic}, and Stable Diffusion \cite{rombach2022high} demonstrated that large-scale diffusion models can generate photorealistic and diverse images from complex textual prompts, leveraging deep language understanding. 
    More recently, new architectures like FLUX.1 \cite{labs2025flux1kontextflowmatching} provide state-of-the-art generation capabilities.
    While these models provide a strong foundation for image synthesis, they are inherently limited in their ability to render specific, user-provided subjects with high fidelity, as they are trained on broad, generic datasets. This limitation spurred the development of personalization techniques.
    
    \subsection{Text-to-Image Personalization via Fine-tuning}
    \textbf{Dreambooth and Follow-up works.}
    To overcome the challenge of subject-specific generation, fine-tuning emerged as the dominant paradigm. Seminal work of DreamBooth \cite{ruiz2023dreambooth} proposed fine-tuning the full diffusion model on subject images demonstrating promising personalization results.
    Many follow-up works \cite{ram2025dreamblend, marjit2025diffusekrona} were proposed to further improve the personalization performance. 
    However, these approaches are still resource-intensive. 
    In response, parameter-efficient techniques such as LoRA \cite{hu2022lora} became prominent.
    Building on this, methods like CustomDiffusion \cite{kumari2022customdiffusion} and SVDiff \cite{han2023svdiff} further optimized the process by targeting specific components of the diffusion model, such as the cross-attention layers, to reduce memory and storage costs.
    
    \vspace{0.5em}
    
    \noindent\textbf{Textual Inversion and Disentanglement.}
    Alternative approaches focus on manipulating the textual conditioning space \cite{gal2022image}, \cite{weili2024infusion}. Textual Inversion \cite{gal2022textual} learns a new token embedding to represent the subject. Other works, such as DreamArtist \cite{dreamartist2025} and StyleDrop \cite{sohn2023styledrop}, concentrate on disentangling subject identity from style. 
    
    \vspace{0.5em}

    \subsection{Fast Text-to-Image Personalization}
    Precursors to our work, like HyperDreamBooth \cite{ruiz2023hyperdreambooth}, introduced hypernetworks to predict personalized weights of a diffusion model in a two stage process, where the first stage requires 50 days of compute on Nvidia RTX 3090 GPU just to prepare the training data and a second stage of fast finetuning the model requiring 20 sec per subject, both of the dependencies our method eliminates. \citet{hedlin2025hypernet} proposed to train the hypernetwork to mimic the optimization trajectory of finetuning. This approach alleviated the need of paired dataset of image and finetuned-weights of the diffusion model to train the hypernetwork, thus simplifying the training.
    However, both of these approaches require finetuning to achieve satisfactory performance.
    
    {
    Besides hypernetworks, a plethora of methods have been proposed for fine-tuning-free personalization. These include adapter-based approaches \cite{ye2023ip, huang2025patchdpo, wang2025ms, huang2025patchdpo}, encoder-driven methods \cite{li2023blip, wei2023elite, zhang2024ssr, ma2024subject, shi2023instantbooth}, in-painting based \cite{zeng2024jedi, kang2025flux}, and other approaches \cite{patel2024lambda, rout2024semantic}.
    These systems typically introduce lightweight networks (adapters or encoders) trained to condition the diffusion model with subject information provided through images or learned embeddings. While they significantly reduce inference latency and memory requirements, existing methods sometimes struggle to achieve acceptable subject fidelity. 
    As shown in Table~\ref{tab:ft_existing_table}, even small amounts of test-time fine-tuning ($\approx$ 100 steps) can lead to substantial improvements in subject fidelity metrics like CLIP-I and DINO similarity for many methods, suggesting that they fall short in generalizing to unseen subjects without further optimization.
    }
    \vspace{0.5em}
    
    
    Our work aims for both speed and quality, where we uniquely employ an end-to-end trained hypernetwork to predict high-fidelity LoRA weights directly, aiming to match the quality of fine-tuning methods while retaining the speed of encoder-based approaches.
    
    \section{Background and Motivation}
    
    \subsection{Fine-tuning Based Personalization}
    
    Personalizing text-to-image (T2I) diffusion models initially relied heavily on fine-tuning techniques. DreamBooth \cite{ruiz2023dreambooth}, one of the earliest and most widely adopted methods, fine-tunes the entire model or its components using a small number of images for a specific subject. Given a set of subject images and prompts $(\x^{(n)}, \bm c^{(n)})_{n=1}^N$, where $\x^{(n)}$ represents the $n$th subject image and ${\bm c}^{(n)}$ being the prompt describing the image, the diffusion fine-tuning objective is as follows:
    \begin{align}\label{eq:ft_obj}
        {\minimize_{\bm \theta}}~ \mathcal{L}_{\rm FT} := \frac{1}{N} \sum_{n=1}^N \mathbb{E}_{t, {\bm \epsilon}, {\x}_t^{(n)}} \left\| {\bm \epsilon}_{\bm \theta}(\x_t^{(n)}, {\bm c}^{(n)}, t) - {\bm \epsilon} \right\|_{2}^{2}
    \end{align}
    where $\bm \theta$ is the set of parameters of the diffusion model, $\bm \epsilon_{\bm \theta}$ is the diffusion denoiser,  $\x_t^{(n)} = \alpha_t \x_t + \sigma_t \epsilon$ is the noise-added image and $\alpha_t, \sigma_t$ are scalars based on noise scheduler parameters. Dreambooth \cite{ruiz2023dreambooth} and many follow-up works also use regularization to prevent the text-to-image model from overfitting to the given image, text pairs. 
    For that purpose, they often use generic images, of the same class as the subject, generated by the base diffusion model.
    The regularization objective is still the same diffusion objective 
    \begin{align}\label{eq:ft_obj}
        {\minimize_{\bm \theta}}~\mathcal{L}_{\rm reg} := \frac{1}{M}\sum_{i=1}^M \mathbb{E}_{t, {\bm \epsilon}, \widehat{\x}_t^{(i)}} \left\| {\bm \epsilon}_{\bm \theta}(\widehat{\x}_t^{(i)}, \widehat{\bm c}^{(i)}, t) - {\bm \epsilon} \right\|_{2}^{2},
    \end{align}
    where $\widehat{\x}^{(m)}, \widehat{c}^{(m)}$ represent the image, prompt pair for regularization (e.g., images different from the subject's but of the same class as the subject). Often a special rare token, denoted here by [V], is used to distinguish the subject from the class images when regularization is used \cite{ruiz2023dreambooth} (e.g., prompt for the subject image: ``a [V] cat'', and prompt for the regularization images: ``a cat'').  
    The total objective $\minimize_{\bm \theta} ~ \cL_{\rm FT} + \gamma \cL_{\rm reg}$ balances subject reconstruction with regularization from class-based data, allowing the model to preserve prompt compliance while injecting subject identity.
    
    These methods achieve strong subject fidelity, while preserving prompt alignment. However, the main limitation is the computational cost. Each new subject instance demands a separate fine-tuning pass, often requiring 3–5 minutes on high-end GPUs, which makes these approaches unsuitable for real-time or large-scale personalization.

    \begin{table}[t]
    \centering
    \caption{\textbf{Effect of Fine-tuning different methods on Dreambench dataset.} *Results reported in respective papers. $^\dagger$Results reproduced here.}
    \resizebox{0.8\linewidth}{!}{
    \begin{tabular}{l|ccc}
    \toprule
    \textbf{Method} & \textbf{CLIP-I} & \textbf{DINO} & \textbf{CLIP-T} \\
    \midrule
    BLIP-Diffusion + FT* & 0.805 & 0.670 & 0.302 \\
    BLIP-Diffusion \cite{li2023blip} & 0.779 & 0.594 & 0.300 \\
    \midrule
    $\lambda$-ECLIPSE + FT* & 0.796 & 0.682 & 0.304 \\
    $\lambda$-ECLIPSE* \cite{patel2024lambda} & 0.783 & 0.613 & 0.307 \\
    \midrule
    MS-Diffusion + FT* & 0.805 & 0.702 & 0.313 \\
    MS-Diffusion* \cite{wang2025ms}& 0.792 & 0.671 & 0.321 \\
    \midrule
    IP-Adapter Plus + FT $^\dagger$ & 0.832 & 0.718 & 0.301 \\
    IP-Adapter Plus $^\dagger$ & 0.825 & 0.693 & 0.307 \\
    \bottomrule
    \end{tabular}
    }
    \label{tab:ft_existing_table}
    \vspace{-3mm}
    \end{table}
    
    \subsection{Hypernetworks for efficient personalization}
    
    Hypernetworks aim to directly predict the fine-tuned parameters of the diffusion model.
    More specifically, a hypernetwork $\h_{\bm \phi}(\x)$ is parameterized by $\bm \phi$ that maps an input image $\x$ to the parameters $\h_{\bm \phi}(\x) = \bm \theta$. 
    The parameters $\bm \theta$ corresponds to the weights of the denoiser.
    A practical challenge in realizing this is that the output dimension of $\h_{\bm \phi}$ must match the dimension of the parameters $\bm \theta$. 
    This can lead to prohibitively large dimension of $\bm \phi$ for deep neural networks. 
    To remedy this, existing works often only estimate the \textit{low rank adapter} (LoRA) parameters (say $\bm \beta$), which tend to be much more manageable. 
    The hypernetwork consists of an image encoder (typically a frozen ViT~\cite{gal2022image} image-encoder) and a trainable transformer decoder network  that takes image features and outputs $\bm \beta$. 
    Existing literature have proposed many designs for the architecture of hypernetworks $\h_{\bm \phi}$ \cite{ruiz2023dreambooth, hedlin2025hypernet}. We use the lightweight architecture of \cite{ruiz2023dreambooth}.
    
    In the next section, we describe our approach of training the hypernetwork in an end-to-end manner, which requires only input subject images for training and does not require fine-tuning during inference time.
    
    \section{Proposed Method}
    
    \subsection{End to end training of Hypernet}\label{sec:end-to-end}
    Consider a set of subject images $\x = \{\x^{(1)}, \dots, \x^{(N)}\}$ for a given subject and finetuning objective $\cL_{\rm FT}$. Finetuning based personalization aims to find  diffusion model parameters $\bm \theta^{\star}_{\x}$ using the following criterion:
    \begin{align}\label{eq:ft_optimal}
        \bm \theta^{\star}_{\x} = \arg \min_{\bm \theta} ~~ \cL_{\rm FT} (\x; \bm \theta) + \gamma \cL_{\rm reg} (\bm \theta).
    \end{align}
    In the context of personalization, hypernetworks are auxiliary neural networks that seek to amortize the cost of finetuning per subject by directly predicting the finetuned parameters $\theta_{\x}^{\star}$ for subject image set $\x$. 
    Let $\bm \theta_0$ denote the base diffusion model parameters before fine-tuning. 
    For the efficiency of optimization, the hypernetwork $\h_{\bm \phi}(\x))$ predicts LoRA parameters for $\bm \theta_0$.
    Then the objective of hypernetwork is to 
    \begin{align}\label{eq:hypernet_lora}
        \minimize_{\bm \phi} &~~ \cL_{\rm FT} (\x; (\bm \theta_0, \h_{\bm \phi}(\x))) + \gamma \cL_{\rm reg} ((\bm \theta_0, \h_{\bm \phi}(\x)))
    \end{align}
    Here, the goal of hypernet is to make the optimal solution $\bm \theta^{\star}$ of Problem \eqref{eq:ft_optimal} satisfy $\h_{\bm \phi^{\star}}(\x) = \bm \beta_{\x}^{\star}$.

    \begin{figure}
        \centering
        \includegraphics[width=0.75\linewidth]{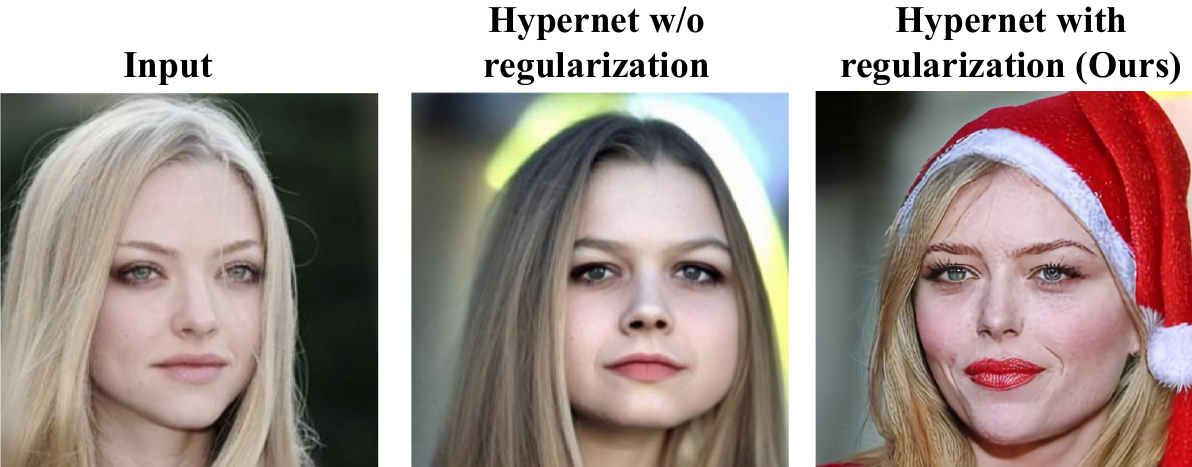}
        \caption{\textbf{Effect of regularization.} The hypernetwork trained without regularization results in poor prompt alignment due to overfitting. Proposed regularization fixes the issue (Sec.~\ref{sec:end-to-end}). Prompt: ``a person wearing a santa hat''.}
        \label{fig:naive_hypernet}
        \vspace{-3mm}
    \end{figure}
    \begin{figure}[t]
        \centering
        \includegraphics[width=0.8\linewidth]{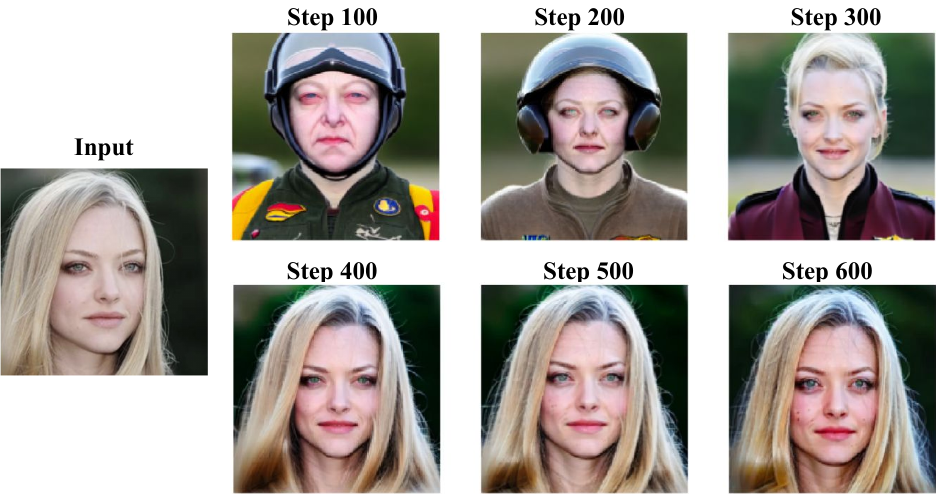}
        \caption{\textbf{Result of Dreambooth Finetuning} at different steps for the prompt ``a person as a top gun pilot''. Early stopping is important to prevent overfitting to input subject image.}
        \label{fig:db_finetuning}
        \vspace{-3mm}
    \end{figure}
    
    Although hypernetworks offer a compelling direction for eliminating per-subject finetuning by predicting finetuned parameters directly, naively optimizing the objective in \eqref{eq:hypernet_lora} does not yield satisfactory results in practice (see Fig. \ref{fig:naive_hypernet}). Specifically, prompt alignment suffers severely. This has also been observed in existing works \cite{hedlin2025hypernet}. 
    The reason for the failure can be traced back to the fact that finetuning loss $\cL_{\rm FT} + \gamma \cL_{\rm reg}$ in \eqref{eq:ft_obj} requires early stopping to prevent overfitting \cite{ruiz2023dreambooth, ram2025dreamblend}.
    This is evident in Fig. \ref{fig:db_finetuning}, where we show the result of Dreambooth finetuning at different steps for the prompt ``a person as a top gun pilot''. As the optimization progresses, we observe that subject fidelity improves but prompt fidelity declines, highlighting a critical need for early stopping during finetuning.
    
    This insight presents a fundamental mismatch: while finetuning-based methods such as Dreambooth can benefit from early stopping by runninng limited number of gradient steps, the end-to-end hypernetwork formulation in \eqref{eq:hypernet_lora} lacks a direct mechanism to control or emulate early stopping because limiting the number of gradient descent steps for parameter $\bm \phi$ does not directly translate to any gradient based early stopping for the output of hypernet $\h_{\bm \phi}(\x)$. 
    
    
    Interestingly, our study reveals that a simple $\ell_2$ regularization on the output of the hypernetwork often suffices to resolve this issue. By penalizing the norm of the predicted LoRA weights, we control the size of the change in LoRA parameters $\bm \beta$,  effectively mimicking the behavior of early-stopped solutions. The proposed objective is as follows:
    \begin{align}\label{eq:hypernet_lora_reg}
        \minimize_{\bm \phi} &~~ \cL_{\rm FT} (\x; (\bm \theta_0, \h_{\bm \phi}(\x))) + \gamma \cL_{\rm reg} ((\bm \theta_0, \h_{\bm \phi}(\x))) \nonumber \\
        & ~~+ \lambda \|\h_{\bm \phi}(\x))\|_2^2
    \end{align}
    Empirical results in Sec. \ref{ref:experiments} demonstrate that adding the regularization in \eqref{eq:hypernet_lora_reg} leads to performance that often exceeds that of more elaborate methods.

    \subsection{Hybrid Model Classifier Free Guidance}\label{sec:hm-cfg}
    Training objectives for most of the existing personalization methods (both fine-tuning-based and fine-tuning-free) are prone to overfitting, even when strong regularization is applied. We observed that this issue is particularly severe in small models such as Stable Diffusion v1.5 (SD1.5), where overfitting to the subject image leads to poor generalization and degraded prompt fidelity. We demonstrate in Sec. \ref{ref:experiments} Fig.~\ref{fig:hmg_qual} that in some cases with complex prompt, the proposed method can also result in weak alignment with prompts. 
    
    To address this, we propose a general inference scheme called {Hybrid Model Classifier Free Guidance} (HM-CFG) that exploits the compositional nature of score function in diffusion models \cite{liu2022compositional} to combine the strengths of both the base and fine-tuned diffusion models during inference. This method is, in principle, applicable to any of the existing methods as a drop-in replacement for CFG-based sampling. The core intuition is as follows:
    
    \begin{itemize}
        \item The base diffusion model excels at prompt understanding and diverse image generation but lacks subject-specific detail.
        \item The fine-tuned (e.g., Hypernet) model captures the subject identity well but often overfits (e.g., produces the same input subject images regardless of the prompt).
    \end{itemize}
    
    HM-CFG combines the strengths of the two models enabling a smooth tradeoff between prompt and subject fidelity. To clarify, consider two prompts: $\bm c_{\rm S}$, representing the subject iden specific prompt (e.g., ``a [V] face as a Minecraft character''), and $\bm c_{\rm G}$, a generic prompt representing the desired composition (e.g., ``a face as a Minecraft character''). Our goal is to sample image $\x$ that is consistent with both prompts $\bm c_{\rm S}$ and $\bm c_{\rm G}$. To that end, let $p(\x_t)$ is the true noise-added image distribution and $\bm c = \{\bm c_{\rm S}, \bm c_{\rm G}\}$. The success of Classifier Guidance~\cite{dhariwal2021diffusion} and CFG~\cite{ho2022classifier} relies on the idea of boosting/guiding the score function $ \s(\x_t, \bm c) := \nabla_{\x_t} \log p(\x_t | \bm c)$ by using the classifier's score $\nabla_{\x_t} \log p(\bm c | \x_t)$, i.e. using the $\widetilde{\s}(\x_t, \bm c) := \nabla_{\x_t} \log p(\x_t | \bm c) + w \log p (\bm c \mid \x_t)$ instead of just $\s(\x_t, \bm c)$, where $w$ is the guidance strength. Using this score in the reverse diffusion process was observed to be equivalent to generating approximate samples from $\widetilde{p}(\x_t \mid \bm c) \propto p(\x_t \mid \bm c) p(\bm c \mid \x_t)^w \propto p(\x_t)  p(\bm c | \x_t)^{w + 1}$. Thus the score $\widetilde{s}(\x_t, \bm c)$ used in CFG is given by \cite{ho2022classifier}:
    \begin{align}\label{eq:cfg}
      \widetilde{s}(\x_t, \bm c) &= \nabla_{\x_t} \log \widetilde{p}(\x_t \mid \bm c) \\
     &=  \nabla_{\x_t}\log p(\x_t) + (w + 1) \nabla_{\x_t} \log p(\bm c | \x_t).\nonumber
    \end{align}
    
    CFG uses the fact that $p(\bm c | \x_t) \propto p(\x_t | \bm c) / p(\x_t)$ to further factorize the classifier score $\nabla_{\x_t} \log p(\bm c | \x_t)$ as $\nabla_{\x_t} \log p(\x_t | \bm c) - \nabla_{\x_t} \log p(\x_t)$. In similar spirit, in HM-CFG, we use the factorization $p(\bm c | \x_t) = p(\bm c_{\rm S}, \bm c_{\rm G} | \x_t) =  p(\bm c_{\rm S}| \x_t) p(\bm c_{\rm G} | \x_t)$ (true for conditionally independent $c_1$ and $c_2$ given $\x_t$) and obtain $p(\bm c | \x_t) = p(\bm c_{\rm S} | \x_t) p(\bm c_{\rm G} | \x_t) \propto p(\x_t | \bm c_{\rm S}) p(\x_t | \bm c_{\rm G}) / {p(\x_t)^2}$. 
    Hence, we obtain the following score function (see Appendix for the complete derivation):
    \begin{align}
    &\nabla_{\x_t} \log \widetilde{\bm \epsilon}(\x_t \mid \bm c_{\rm S}, \bm c_{\rm G}) = \nabla_{\x_t} \log p(\x_t) + (w + 1) \times \\
    & \left( \nabla_{\x_t} \log p(\x_t \mid \bm c_{\rm S}) + \nabla_{\x_t} \log p(\x_t \mid \bm c_{\rm G}) - 2\nabla_{\x_t} \log p(\x_t) \right) \nonumber
    \end{align}
    Here, all the scores are based on the true data distribution. Generally, during inference via CFG in \eqref{eq:cfg}, (personalized) diffusion models replace both conditional and unconditional scores, $\nabla_{\x_t} \log p(\x_t \mid \bm c)$ and $\nabla_{\x_t} \log p(\x_t)$, via the scaled denoiser $\bm \epsilon_{\bm \theta}( \x_t, \bm c) / \sigma_t$ and $\bm \epsilon_{\bm \theta}( \x_t, \varnothing) / \sigma_t$, where $\sigma_t$ is a scalar determined by the noise schedule, $\theta$ is the optimized (personalized) diffusion model parameters, and  $\varnothing$ represents the empty prompt. 
    
    However, 
    $\bm{\epsilon}_{\bm{\theta}}$ often overfits to the subject images, which compromises prompt following and reduces image diversity compared to the base diffusion model $\bm{\epsilon}_{\bm{\theta_0}}$. As a result, $\bm{\epsilon}_{\bm{\theta_0}}$ yields superior estimates for both the unconditional score $\nabla_{\mathbf{x}_t} \log p(\mathbf{x}_t)$ and the generic prompt conditional score $\nabla_{\mathbf{x}_t} \log p(\mathbf{x}_t \mid \mathbf{c}_{\rm G})$, owing to its inherent diversity and capacity for generic prompt following, respectively. In contrast, $\bm{\epsilon}_{\bm{\theta}}$ more accurately estimates subject-specific prompt conditional score $\nabla_{\mathbf{x}_t} \log p(\mathbf{x}_t \mid \mathbf{c}_{\rm S})$ by leveraging its specific subject knowledge. Hence HM-CFG uses the following expression to model the effective noise estimate at each denoising step:
    \begin{align*}
        &\widetilde{\bm \epsilon} (\x_t, \bm c) = {\bm \epsilon}_{\bm \theta_0}(\x_t, \varnothing) + (w + 1) \times \\
    & \left({\bm \epsilon}_{\bm \theta}(\x_t, \bm c_{\rm S}) + {\bm \epsilon}_{\bm \theta_0}(\x_t, \bm c_{\rm G}) - 2{\bm \epsilon}_{\bm \theta_0, \varnothing}(\x_t) \right) 
    \vspace{-1mm}
    \end{align*}
    
    This combination allows us to combine the strength of high subject fidelity from the fine-tuned model while preserving prompt alignment and visual diversity from the base model. 
    HM-CFG also allows us to trade-off between prompt and subject fidelity by simply using a weighting factor $\kappa \in [0,2]$ as follows:
    \begin{align*}
        &\widetilde{\bm \epsilon} (\x_t, \bm c) = {\bm \epsilon}_{\bm \theta_0}(\x_t, \varnothing) + (w + 1) \times \\
    & \left( \kappa {\bm \epsilon}_{\bm \theta}(\x_t, \bm c_{\rm S}) + (2-\kappa){\bm \epsilon}_{\bm \theta_0}(\x_t, \bm c_{\rm G}) - 2{\bm \epsilon}_{\bm \theta_0}(\x_t, \varnothing) \right). 
    \end{align*}
    
    \begin{figure}[t]
        \centering
        \includegraphics[width=0.8\linewidth]{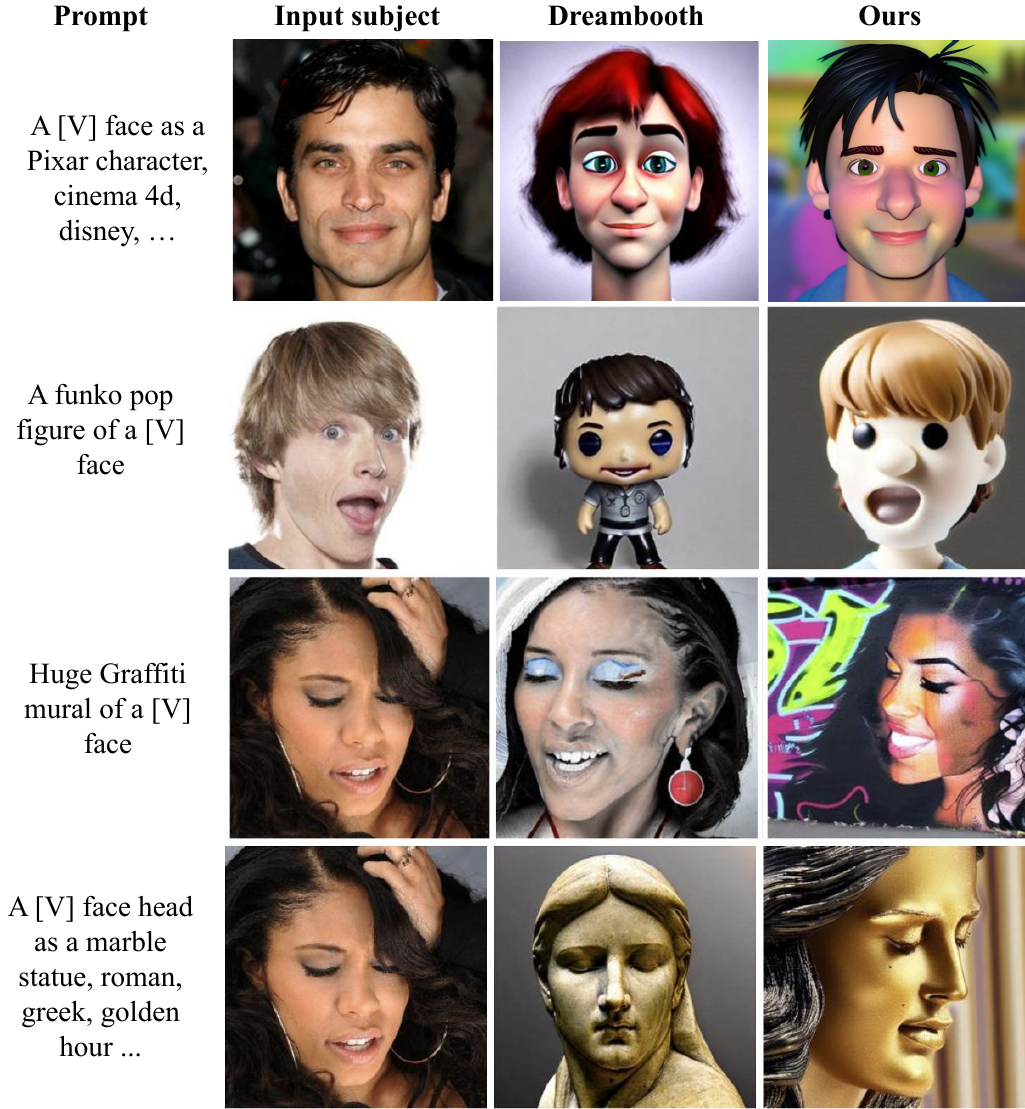}
        \caption{\textbf{Qualitative results on CelebA-HQ dataset.} Proposed method shows competitive subject and prompt fidelity compared to fine-tuning-based approach Dreambooth.}
        \label{fig:celeba_qualitative_results}
        \vspace{-2mm}
    \end{figure}
    
    \section{Experiments}\label{ref:experiments}
    We now evaluate our proposed hypernetwork framework on both closed-category and open-category personalization tasks. We demonstrate that our method achieves state-of-the-art performance among existing approaches: hypernetwork based solutions and other fine-tuning-free approaches.
    All evaluation results reported on Sec. \ref{sec:exp_closed_cat} and \ref{sec:exp_open_cat} are obtained by using CFG-based inference for a fair comparison with all the baselines which also used CFG for inference. The results of HM-CFG are in the Sec. \ref{sec:exp_hm_cfg} as well as the Appendix.
    
    \noindent\textbf{Metrics}
    We use widely adopted metrics \cite{ruiz2023dreambooth, huang2025patchdpo} based on DINO \cite{caron2021emerging} and CLIP \cite{radford2021learning} feature similarity of generated and input images for measuring the subject image alignment. To measure the prompt alignment, we use CLIP feature similarity between the generated images and the prompt applied for generation. We use \texttt{CLIP-I} and \texttt{CLIP-T} to refer to the CLIP image and text alignment, respectively.
    
    \subsection{Closed-Category Personalization}\label{sec:exp_closed_cat}
    \noindent\textbf{Datasets.}
    We use the same benchmark datasets as used by \cite{hedlin2025hypernet}: CelebA-HQ \cite{karras2017progressive} and AFHQ-v2 \cite{choi2020stargan}. 
    We follow the same train/test split and testing prompts as in \cite{hedlin2025hypernet}. For CelebA-HQ, we use 29,800 images for training and 100 images for evaluation, whereas for AFHQ-v2, we use randomly selected 100 images from its test set for evaluation and all images in the training set for training. 
    For the CelebA-HQ dataset, we also employ \texttt{Facerec.} as another metric that measures the face embedding similarity between the generated and input images based on VGGFace2 \cite{cao2018vggface2}.
    
    \noindent\textbf{Baselines.}
    For the closed-category tasks, our primary baseline is the recent work of \cite{hedlin2025hypernet} referred to as \texttt{HyperNetField}. We also use \texttt{HyperNetField + FT} as a baseline, which uses 50 steps of per-subject finetuning at test time. However, we will see that our method does not require per-subject finetuning to attain the same performance. Besides, we also use finetuning-based method \texttt{Dreambooth} \cite{ruiz2023dreambooth} as a baseline.
    
    \noindent\textbf{Settings.}
    For a fair comparison with baselines, we also employ the same diffusion model \textit{Stable Diffusion v1.5} (SD1.5) \cite{rombach2022high} as our base model. The hypernetwork is trained to predict LoRA parameters for all the cross-attention layers of the diffusion model's U-Net. We use a rank of 3 for the LoRA matrices, which results in 223.5k total number of LoRA weights. We use the same hypernetwork architecture as in \cite{ruiz2023hyperdreambooth}, i.e., ViT base image encoder \cite{wu2020visual} pre-trained on Imagenet 21k \cite{deng2009imagenet}. We use Adam optimizer with an initial learning rate of $1e-5$ and total batchsize of 64 distributed across 4 H100 GPUs. Details of the hypernet architecture are in the Appendix.
    
    \noindent\textbf{Results.}
    Table \ref{tab:afhq_results} presents results attained by all methods on the AFHQ dataset, which contains non-human subjects (e.g., cats, dogs, and wild animals). Our method with $\lambda=0.15$ consistently outperforms \texttt{Dreambooth} and \texttt{HyperNetField} with per-subject finetuning. Note that our method does not use any finetuning. Qualitative results on AFHQ-v2 dataset are shown in the Appendix.
    
    Table \ref{tab:celeba_results} shows the results attained by all methods on CelebA-HQ dataset. One can see that the proposed method with regularization constant $\lambda=0.15$ achieves the best subject alignment (as measured by \texttt{DINO} and \texttt{CLIP-I}) while maintaining comparable prompt alignment with the baselines. This is a substantial improvement over \texttt{DreamBooth}, which needs approximately 180 seconds per subject, and \texttt{HyperNetField}, which still requires 20 seconds of test-time fine-tuning to be effective. 
    The results underscore the importance of our proposed regularization on the output of hypernetwork; the unregularized version of our model ($\lambda=0.0$) overfits significantly, evidenced by a sharp drop in prompt fidelity (\texttt{CLIP-T}: 0.226) despite inflated subject similarity scores. Fig. \ref{fig:celeba_qualitative_results} shows some qualitative results on the CelebA-HQ dataset. We can see that the results are comparable (if not better) than full finetuning based approach.

    
    \begin{table}[t]
    \centering
    \caption{\textbf{Performance comparison of different hypernetwork types on the AFHQ-v2 dataset}.  *Results from \cite{hedlin2025hypernet}.}
    \resizebox{\linewidth}{!}{
    \begin{tabular}{l|cccc}
    \toprule
    \textbf{Hypernet Type} & \textbf{DINO} & \textbf{CLIP-I} & \textbf{CLIP-T} & \textbf{FT time (s)}\\
    \midrule
    Dreambooth (LoRA)* & 0.560 & 0.763 & 0.268 & $\approx$180\\
    HyperNetField + FT* & 0.664 & 0.807 & 0.277 & $\approx20$\\
    HyperNetField* & 0.495 & 0.746 & \textbf{0.285} & 0\\
    Ours $(\lambda=0.5)$ & \textbf{0.717} & \textbf{0.813} & 0.278 & \textbf{0}\\
    \bottomrule
    \end{tabular}}
    \label{tab:afhq_results}
    \vspace{-1mm}
    \end{table}

    \begin{table}[t]
    \centering
    \caption{\textbf{Results on the CelebA-HQ dataset.} *Results from \cite{hedlin2025hypernet}. $^\dagger$ Result from \cite{ruiz2023hyperdreambooth}.}
    \resizebox{\linewidth}{!}{
    \begin{tabular}{l|ccccc}
    \toprule
    \textbf{Method} & \textbf{DINO} $\uparrow$ & \textbf{CLIP-I} $\uparrow$ & \textbf{Face Rec.} $\uparrow$ & \textbf{CLIP-T} $\uparrow$ & \textbf{FT time (s)} \\
    \midrule
    Dreambooth (CFG) & 0.539 & 0.609 & 0.356 & 0.275 & $\approx$180 \\
    Hyperdreambooth + FT $^\dagger$ &  0.473 & 0.577 & \textbf{0.655} & \textbf{0.286} & $\approx$ 20 \\
    HyperNetField + FT* & 0.605 & 0.639 & 0.325 & 0.268 & $\approx$20 \\
    HyperNetField* & 0.532 & 0.582 & 0.157 & 0.284 & \textbf{0} \\
    Ours ($\lambda=0.15$) & \textbf{0.639} & \textbf{0.653} & 0.250 & 0.269 & \textbf{0} \\
    \midrule
    Ours ($\lambda=0.0$) & 0.723 & 0.706 & 0.265 & 0.226 & 0 \\
    \bottomrule
    \end{tabular}
    }
    \label{tab:celeba_results}
    \vspace{-1mm}
    \end{table}

    \subsection{Open-Category Personalization}\label{sec:exp_open_cat}
    
    \begin{table}[t!]
    \centering
    \caption{\textbf{Comparison of different methods on Dreambench.} ``Knd.'' is short for Kandinsky v2.2, ``Inv.'' for Inversion and ``Diff.'' for Diffusion. $^\dagger$ Results reported from their respective papers. *Result reported in \cite{li2023blip}. Other results are reproduced here.}
    \resizebox{\linewidth}{!}{
    \begin{tabular}{l|ccccc}
    \toprule
    \textbf{Method} & Model & \textbf{DINO} & \textbf{CLIP-I} & \textbf{CLIP-T} & \textbf{Avg.} \\
    \midrule
    Real Images* & -- & 0.774 & 0.885 & N/A & N/A \\
    \midrule
    Textual Inv. $^\dagger$ & SD1.5 &  0.569 & 0.780 & 0.255 & 0.535 \\
    DreamBooth $^\dagger$ & Imagen & 0.696 & 0.812 & 0.306 & 0.605 \\
    DreamBooth $^\dagger$ & SD1.5 & 0.668 & 0.803 & 0.305 & 0.592 \\
    Custom Diff. $^\dagger$ & SD1.5 & 0.643 & 0.790 & 0.305 & 0.579 \\
    \midrule
    ELITE $^\dagger$ & SD1.4 & 0.652 & 0.762 & 0.255 & 0.556 \\
    SSR-Encoder $^\dagger$ & SD1.5 & 0.612 & 0.821 & {0.308} & 0.580 \\
    BLIP-Diff $^\dagger$. & SD1.5 & 0.594 & 0.779 & 0.300 & 0.558 \\
    Subject-Diff. $^\dagger$ & SD1.5 &  0.711 & 0.787 & 0.293 & 0.597 \\
    JeDi $^\dagger$ & SD1.4 &  0.679 & 0.814 & 0.293 & 0.595 \\
    \midrule
    RF-Inv $^\dagger$. & FLUX & 0.619 & 0.787 & 0.294 & 0.567\\
    LatentUnfold $^\dagger$ &  FLUX & 0.660 & 0.806 & 0.305 & 0.590\\
    OmniControl $^\dagger$ & FLUX & 0.627  & 0.773 & \textbf{0.322} & 0.574 \\
    \midrule
    IP-Adapter-Plus & SDXL & 0.693 & 0.825 & 0.307 & 0.608 \\
    PatchDPO $^\dagger$ & SDXL & 0.727 & 0.838 & 0.292 & 0.619 \\
    Ours ($\lambda=100$) & SDXL & \textbf{0.739} & \textbf{0.846} & 0.293 & \textbf{0.626} \\
    \bottomrule
    \end{tabular}
    }
    \label{tab:open_result}
    \vspace{-3mm}
    \end{table}
    
    \begin{figure}[t]
        \centering
        \includegraphics[width=\linewidth]{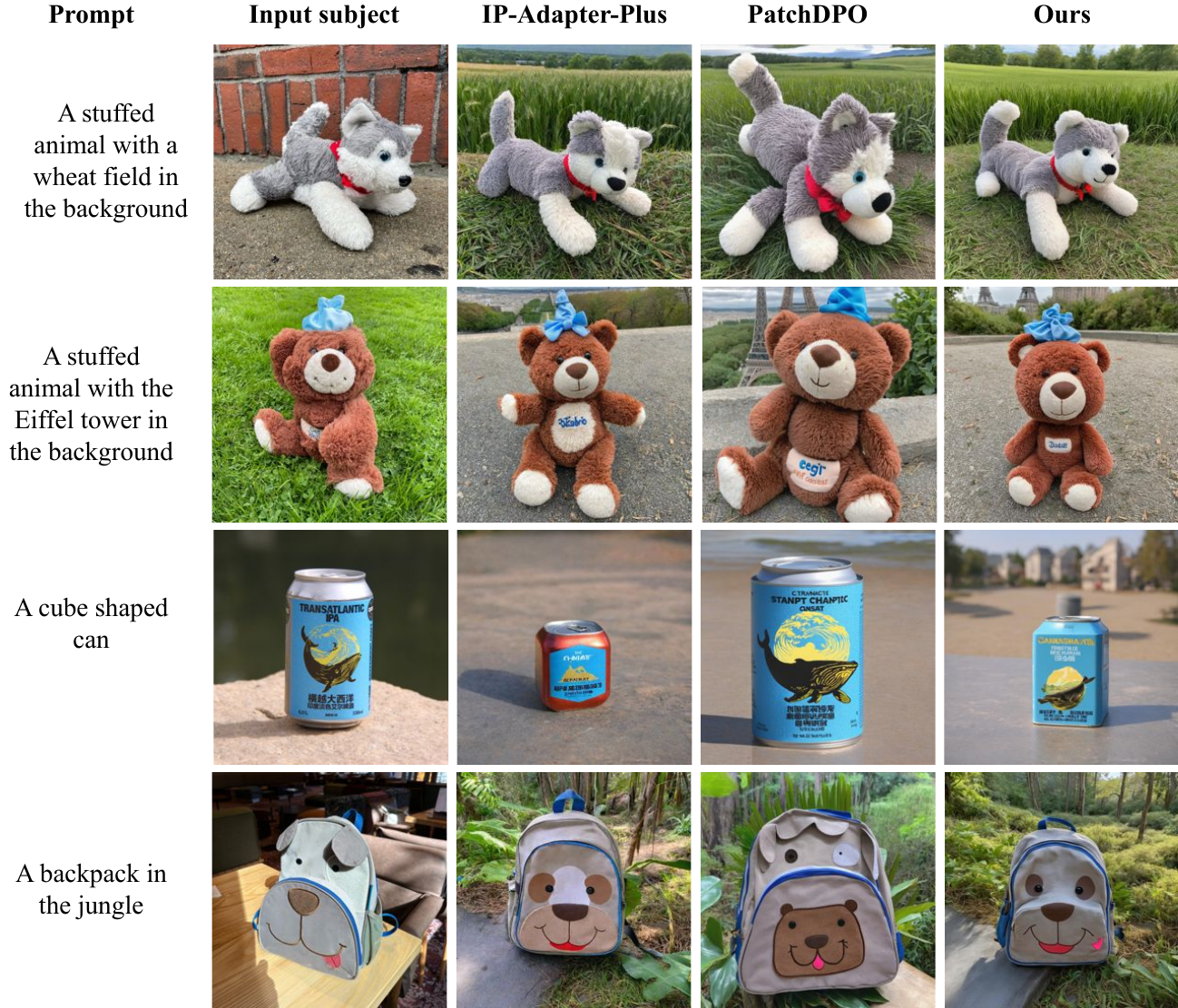}
        \caption{\textbf{Qualitative results on Dreambench dataset.} Notable improvement in subject fidelity and prompt adherence over the baselines can be observed.}
        \label{fig:open_result}
        \vspace{-2mm}
    \end{figure}
    
    \begin{table}[t]
    \centering
    \caption{\textbf{Result of HM-CFG on CelebA-HQ with $\kappa=1$.} Note that HM-CFG is applicable to all methods relying on CFG-based diffusion sampling.}
    \resizebox{\linewidth}{!}{
    \begin{tabular}{l|cccc}
    \toprule
    \textbf{Guidance Type} & \textbf{DINO} $\uparrow$ & \textbf{CLIP-I} $\uparrow$ & \textbf{Face. Rec.} $\uparrow$ & \textbf{CLIP-T} $\uparrow$ \\
    \midrule
    Dreambooth (CFG) & 0.539 & 0.609 & 0.356 & 0.275 \\
    Dreambooth (HM-CFG) & \textbf{0.563} & \textbf{0.632} & \textbf{0.407} & \textbf{0.278} \\
    \midrule
    Ours (CFG) & {0.639} & {0.653} & \textbf{0.250} & 0.269 \\
    
    Ours (HM-CFG)  & \textbf{0.652} & \textbf{0.667} & 0.249 & \textbf{0.275} \\
    \bottomrule
    \end{tabular}
    }
    \label{tab:dual_prompt_results}
    \vspace{-4mm}
    \end{table}

    \noindent\textbf{Datasets.}
    We use publicly available synthetic dataset Subject200k \cite{tan2024omini} for open-category training. The dataset was generated by using Flux.1 dev on LLM-synthesized prompts of single object on diverse backgrounds.
    We evaluate our model and all baselines on  Dreambench \cite{ruiz2023dreambooth}, which is the most widely adopted benchmark for open-category personalization. 
    
    \vspace{1mm} 

    \noindent\textbf{Baselines.}
    We compare our method with a variety of baselines from fine-tuning-based to training-free methods, listed in Table \ref{tab:open_result}, across a variety of base diffusion models. 
    All baseline results are extracted from their respective papers unless otherwise stated. Full description is in the Appendix.
    
    \vspace{1mm} 

    \noindent\textbf{Settings.}
    Training for open-categoy images can be computationally costly given the need for generalization over a vast number of possible subjects. 
    Therefore, it is common to bootstrap from an existing personalization model for computational efficiency \cite{huang2025patchdpo, wang2025ms}. Hence, we also use the popular model IP-Adapter Plus as the base model. 
    We noticed that using IP-Adapter Plus \cite{ipadaptergit} results in much better personalization and faster convergence in comparison to vanilla diffusion model (see Appendix for more details).
    To minimize redundancy, our hypernetwork architecture shares the same CLIP image encoder of IP-Adapter Plus and uses the same resampler network architecture as the LoRA weight decoder.
    Similar to the closed-category setting, we only predict the LoRA parameters of the cross-attention matrices of SDXL. We use $\lambda=100$. We use Adam optimizer with a batch size of 64 divided across 4 H100 GPU and train for 4,000 steps. To leverage multiple images per subject available in Dreambench dataset, we take the average of the output of the hypernet, i.e., $\bm \beta_{\x} = \frac{1}{N} \sum_{n=1}^N \h_{\bm \phi}(\x^{(n)})$.
    
    \vspace{1mm} 
    
    \noindent\textbf{Results.} 
    We present our quantitative results in Table \ref{tab:open_result}. 
     We group methods based on whether they are fine-tuning based (Row 2-5), and base model used (SD1.4/1.5, FLUX, and SDXL). The reason that we have generally observed (also see results in their respective papers) these aspects to significantly impact both qualitative as well as quantitative results. For e.g., methods based on FLUX exhibit superior prompt following and subject detail preservation in general. Hence, fair comparison would generally require using the same base model and fine-tuning steps.
    Nonetheless, our method achieves state-of-the-art performance on the Dreambench benchmark based on the average score. This demonstrates our model's strong ability to maintain subject fidelity. Moreover, it only takes about an hour on the 4 GPUs for the complete training, which is significantly lower than most of the baselines and much lower than the second best baseline \texttt{PatchDPO} which takes about 4 hours on 8 GPUs. 
    
    Figure \ref{fig:open_result} presents qualitative comparisons on the Dreambench dataset.
    To isolate the contribution of each method from that of the base model, we use baselines built on the same backbone (SDXL) without any fine-tuning.
    Under this setting, our approach produces images with higher subject fidelity than IP-Adapter Plus and PatchDPO—for example, more faithfully capturing the details of the stuffed animal and the can while placing them in the correct prompt-specified context.
    Although larger base models such as FLUX or subject-specific post-finetuning can further enhance subject detail and prompt alignment, such results primarily reflect the effect of additional training rather than the intrinsic capability of a method. For this reason, and unlike prior work (e.g., MS-Diffusion \cite{wang2025ms}), we focus on results without post-finetuning

    \begin{figure}[t]
        \centering
        \includegraphics[width=0.75\linewidth]{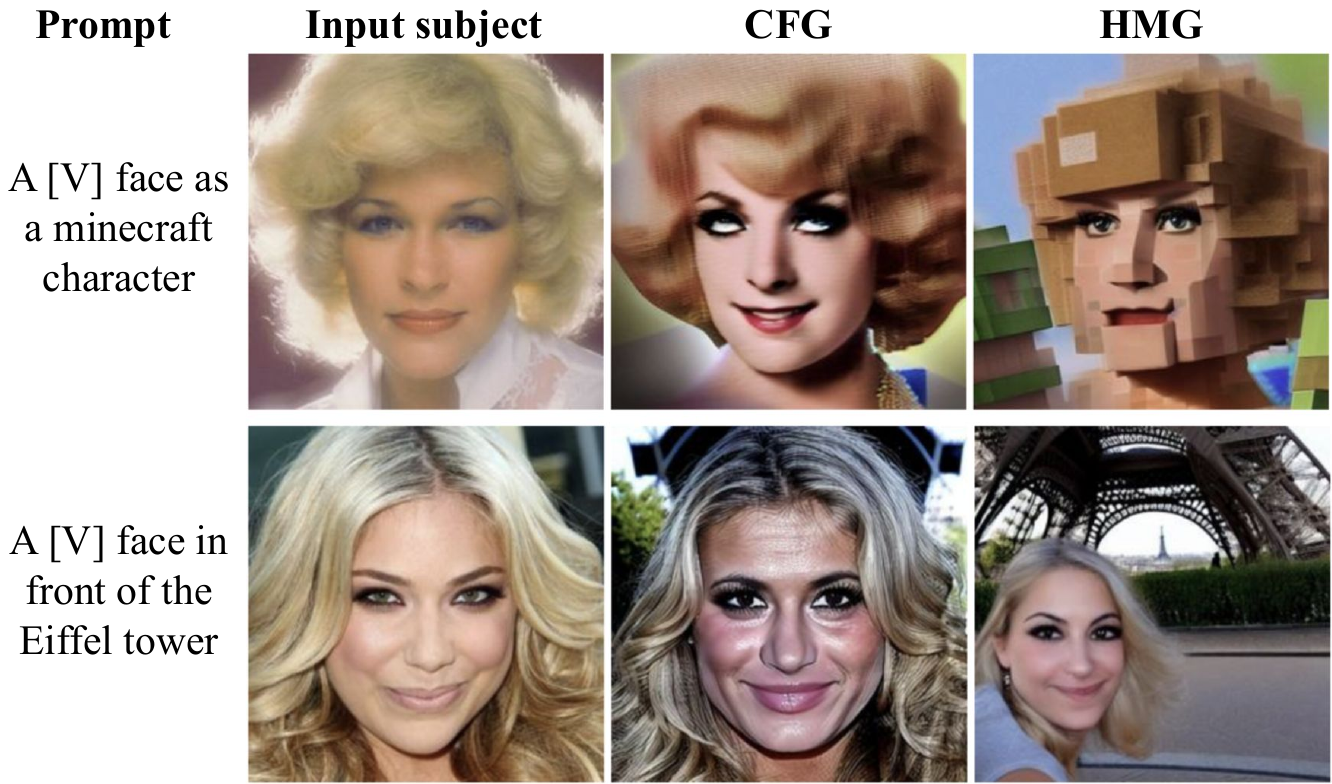}
        \caption{\textbf{Qualitative results of applying HM-CFG for on CelebA-HQ.} Significant improvement in prompt alignment can be observed.}
        \label{fig:hmg_qual}
        \vspace{-3mm}
    \end{figure}

    \subsection{Hybrid-Model CFG}\label{sec:exp_hm_cfg}
    \vspace{-1mm}
    Here, we evaluate the effectiveness of HM-CFG inference technique. Table \ref{tab:dual_prompt_results} shows that this inference-time strategy provides a clear benefit. The results are obtained for $\kappa=1.0$ and $w=5.0$. For DreamBooth, it improves performance across all metrics, boosting subject and prompt fidelity simultaneously. For our HyperNet, it also improves prompt alignment (\texttt{CLIP-T} from 0.269 to 0.275) while keeping the subject fidelity unchanged, demonstrating its value in improving prompt fidelity. More quantitative results, including the potential of controlling subject-prompt trade-off by varying $\kappa$, are in the Appendix.  
    
    Fig. \ref{fig:hmg_qual} shows some qualitative result of HM-CFG on CelebA-HQ dataset. One can see that CFG often fails to follow complex prompts. HFM can significantly improve the prompt following while maintaining subject fidelity. It can also be observed that due to the nature of prompts, minor degradation in subject fidelity is expected and does not undermine the personalization goal. More qualitative results, including the results of HM-CFG on other baselines, are in the Appendix. 
    
    \section{Conclusion}
    In this work, we introduced an end-to-end trained hypernetwork that enables high-fidelity, fine-tuning-free personalization of text-to-image models. By leveraging a simple $\ell_2$ regularization on the predicted LoRA weights to prevent overfitting and a novel inference approach called Hybrid Model CFG for enhanced compositional control at inference, our framework eliminates the need for costly per-subject optimization. Our experiments demonstrate state-of-the-art results across standard benchmarks, including CelebA-HQ, AFHQ-v2, and DreamBench, outperforming existing tuning-free methods while remaining orders of magnitude faster than traditional fine-tuning. Ultimately, our approach offers an efficient solution for producing on-demand, subject-driven generative content.
    
    \bibliography{aaai2026}
    
    \clearpage
    \appendix
    
    \begin{center}
        {\large {\bf Supplementary Material of ``Finetuning-Free Personalization of Text to Image Generation via Hypernetworks''}}
    \end{center}

    \section{Details on HM-CFG}
    
    \subsection{Diffusion Model Sampling}
    Diffusion models generate samples by simulating a reverse stochastic process that transforms noise into data. This is achieved by learning to approximate the reverse of a \textit{forward diffusion process} that gradually adds noise to data.
    
    \paragraph{Forward Process.} The forward process defines a Markov chain:
    \begin{align}
        q(\x_t | \x_{t-1}) = \mathcal{N}(\x_t; \sqrt{1 - \beta_t} \x_{t-1}, \beta_t \mathbf{I}),
    \end{align}
    where $\beta_t$ controls the amount of Gaussian noise added at each step. After $T$ steps, $\x_T$ approximately follows a standard Gaussian distribution: $q(\x_T) \approx \mathcal{N}(\mathbf{0}, \mathbf{I})$.
    
    \paragraph{Reverse Process and the Role of the Score Function.} To sample from the data distribution, we run the reverse process:
    \begin{align}
        p_\theta(\x_{t-1} | \x_t) = \mathcal{N}(\x_{t-1}; \bm \mu_\theta(\x_t, t), \bm \Sigma_t),
    \end{align}
    where $\bm \mu_\theta$ is the conditional mean learned via neural networks. Under the continuous-time formulation \cite{song2020score}, the reverse process is governed by the \textit{score function}:
    \begin{align}
        \s(\x_t) := \nabla_{\x_t} \log p_t(\x_t),
    \end{align}
    i.e., the gradient of the log-density of the noisy data at time $t$. This guides how to denoise $\x_t$ toward regions of high data density.
    
    \paragraph{Sampling via Score-Based SDE.} A general way to sample is by solving the \textit{reverse-time stochastic differential equation} (SDE) or its deterministic counterpart (e.g., the probability flow ODE), both of which require estimating the score function $\s(\x_t)$. In practice, this is learned via training a noise predictor $\bm \epsilon_\theta(\x_t, t)$, which relates to the score as:
    \begin{align}
        \s(\x_t) \approx -\frac{1}{\sigma_t} \bm \epsilon_\theta(\x_t, t),
    \end{align}
    where $\sigma_t$ is the standard deviation of the noise added at time $t$.
    
    \subsubsection*{Classifier-Free Guidance (CFG).}
    To condition generation on prompt $\bm c$, \textit{classifier-free guidance} (CFG) is the most commonly used sampling technique for (personalized) diffusion model, where the conditional score $\nabla_{\x_t} \log p(\x_t | \bm c)$ is boosted by interpolating with the unconditional score $\nabla_{\x_t} \log p(\x_t)$. The guided score is:
    \begin{align}
        \widetilde{\s}(\x_t, \bm c) = \nabla_{\x_t} \log p(\x_t) + (w + 1) \nabla_{\x_t} \log p(\bm c | \x_t),
    \end{align}
    which effectively biases sampling toward images more consistent with the prompt $\bm c$.
    
    In practice, this score is approximated using denoisers trained with and without conditioning, via:
    \begin{align}
        \widetilde{\bm \epsilon}(\x_t, \bm c) = \bm \epsilon_{\bm \theta}(\x_t, \varnothing) + (w+1) \left(\bm \epsilon_{\bm \theta}(\x_t, \bm c) - \bm \epsilon_{\bm \theta}(\x_t, \varnothing) \right).
    \end{align}

    
    
    
    \subsection{Complete Derivation of HM-CFG}
    Given two prompts $\bm c = \{\bm c_{\rm S}, \bm c_{\rm G}\}$, our objective is to sample from a distribution where the generated image is consistent with both the subject-specific prompt $\bm c_{\rm S}$ and the generic prompt $\bm c_{\rm G}$. Using a classifier-free guidance (CFG) approach, we modify the score used in the reverse process.
    
    The guided score under CFG is given by:
    \begin{align}
        \widetilde{\s}(\x_t, \bm c) &= \nabla_{\x_t} \log p(\x_t | \bm c) + w \nabla_{\x_t} \log p(\bm c | \x_t) \nonumber \\
        &= \nabla_{\x_t} \log p(\x_t) + (w+1)\nabla_{\x_t} \log p(\bm c | \x_t). \label{eq:cfg-score}
    \end{align}
    
    For HM-CFG, we assume conditional independence between the subject and generic prompts given $\x_t$, i.e.,
    \begin{align}
        p(\bm c | \x_t) &= p(\bm c_{\rm S}, \bm c_{\rm G} | \x_t) \nonumber \\
                        &= p(\bm c_{\rm S} | \x_t) p(\bm c_{\rm G} | \x_t).
    \end{align}
    
    Now apply Bayes’ rule to each:
    \begin{align}
        p(\bm c_i | \x_t) &\propto \frac{p(\x_t | \bm c_i)}{p(\x_t)}.
    \end{align}
    Hence, the joint becomes:
    \begin{align}
        p(\bm c | \x_t) &\propto \frac{p(\x_t | \bm c_{\rm S})}{p(\x_t)} \cdot \frac{p(\x_t | \bm c_{\rm G})}{p(\x_t)} = \frac{p(\x_t | \bm c_{\rm S}) p(\x_t | \bm c_{\rm G})}{p(\x_t)^2}.
    \end{align}
    
    Taking the log and gradient:
    \begin{align}
        \nabla_{\x_t} \log p(\bm c | \x_t) =& \nabla_{\x_t} \log p(\x_t | \bm c_{\rm S}) + \nabla_{\x_t} \log p(\x_t | \bm c_{\rm G})\nonumber\\
        &- 2 \nabla_{\x_t} \log p(\x_t).
    \end{align}
    
    Plugging this back into Eq.~\eqref{eq:cfg-score}:
    \begin{align}
    \widetilde{\s}(\x_t, \bm c) 
    &= \nabla_{\x_t} \log p(\x_t) + (w+1) \left[ \nabla_{\x_t} \log p(\x_t | \bm c_{\rm S}) \right. \nonumber\\
    &\quad \left. + \nabla_{\x_t} \log p(\x_t | \bm c_{\rm G}) - 2 \nabla_{\x_t} \log p(\x_t) \right]
    \end{align}
    
    Now using the approximation $\nabla_{\x_t} \log p(\x_t | \bm c) \approx -\bm \epsilon_\theta(\x_t, \bm c)/\sigma_t$ and similarly for other terms, we obtain:
    \begin{align}
        \widetilde{\bm \epsilon}(\x_t, \bm c) =& \bm \epsilon_{\bm \theta_0}(\x_t, \varnothing) + (w + 1) \big( \bm \epsilon_{\bm \theta}(\x_t, \bm c_{\rm S}) + \bm \epsilon_{\bm \theta_0}(\x_t, \bm c_{\rm G}) \nonumber\\
        &- 2 \bm \epsilon_{\bm \theta_0}(\x_t, \varnothing) \big).
    \end{align}
    
    To enable tradeoff between prompt and subject fidelity, we introduce the interpolation factor $\kappa \in [0, 2]$, and generalize the formulation to:
    \begin{align*}
        & \widetilde{\bm \epsilon}(\x_t, \bm c) = \bm \epsilon_{\bm \theta_0}(\x_t, \varnothing) + (w + 1) \times \\
        & \big( \kappa \bm \epsilon_{\bm \theta}(\x_t, \bm c_{\rm S}) + (2 - \kappa) \bm \epsilon_{\bm \theta_0}(\x_t, \bm c_{\rm G}) - 2 \bm \epsilon_{\bm \theta_0}(\x_t, \varnothing) \big).
    \end{align*}
    
    This completes the derivation of the Hybrid Model Classifier-Free Guidance (HM-CFG).

    \section{Experimental Details}
    
    \subsection{Closed-Category Personalization}
    \subsubsection{Settings}
    Table \ref{tab:app_closed_hyper} shows the list of the hyperparameter settings for the closed category personalization task for both CelebA-HQ and AFHQv2 datasets. For CelebA-HQ, we train the model for 30k steps, whereas for AFHQv2 only 4k steps appeared to be sufficient. We use the default ODE solver of SD1.5 for sampling from the personalized diffusion model. The hypernetwork architecture is the same for both CelebA-HQ and AFHQv2 experiments. Specifically, we use network architecture proposed in \cite{ruiz2023hyperdreambooth}. The image encoder is frozen ViT-B/16 \cite{dosovitskiy2020image} and the weight decoder is a 4 hidden layer transformer decoder with embedding dimension of 4. As in \cite{ruiz2023hyperdreambooth}, we use 4 self iterations of the decoder, and use a separate linear layer at the output head per each LoRA weight matrix to be predicted. 
    
    \begin{table}[h!]
    \centering
    \caption{\textbf{Hyperparameter settings used in our closed category experiments.}}
    \begin{tabular}{@{}ll@{}}
    \toprule
    \multicolumn{2}{c}{\textbf{Training Settings}} \\
    \midrule
    Learning Rate (LR) & $1 \times 10^{-5}$ \\
    Optimizer & AdamW \\
    Weight decay & $1 \times 10^{-4}$ \\
    Batch size & $64$ \\
    $\lambda$ (regularization) & $100$ \\
    Scheduler & constant with warmup \\
    LR warmup steps & 500 \\
    Number of steps & $4000$ \\
    Number of H100 hours & $\approx$ 1 \\
    LoRA rank & 3 \\
    LoRA target modules & Q, K, V cross attention matrices \\
    \midrule
    \multicolumn{2}{c}{\textbf{Inference Settings}} \\
    \midrule
    Guidance scale $(w + 1)$ & $7.5$ \\
    Number of diffusion steps & $30$ \\
    \bottomrule
    \end{tabular}
    \label{tab:app_closed_hyper}
    \end{table}

    \subsection{Open-Category Personalization}
    
    \subsubsection{Baselines}
    Table \ref{tab:app_baselines} lists all the baselines used in this work and whether they are fine-tuning based methods. The baselines were picked based on their recency, or relevance, or representativeness.
    
    \begin{figure}[t]
        \centering
        \includegraphics[width=\linewidth]{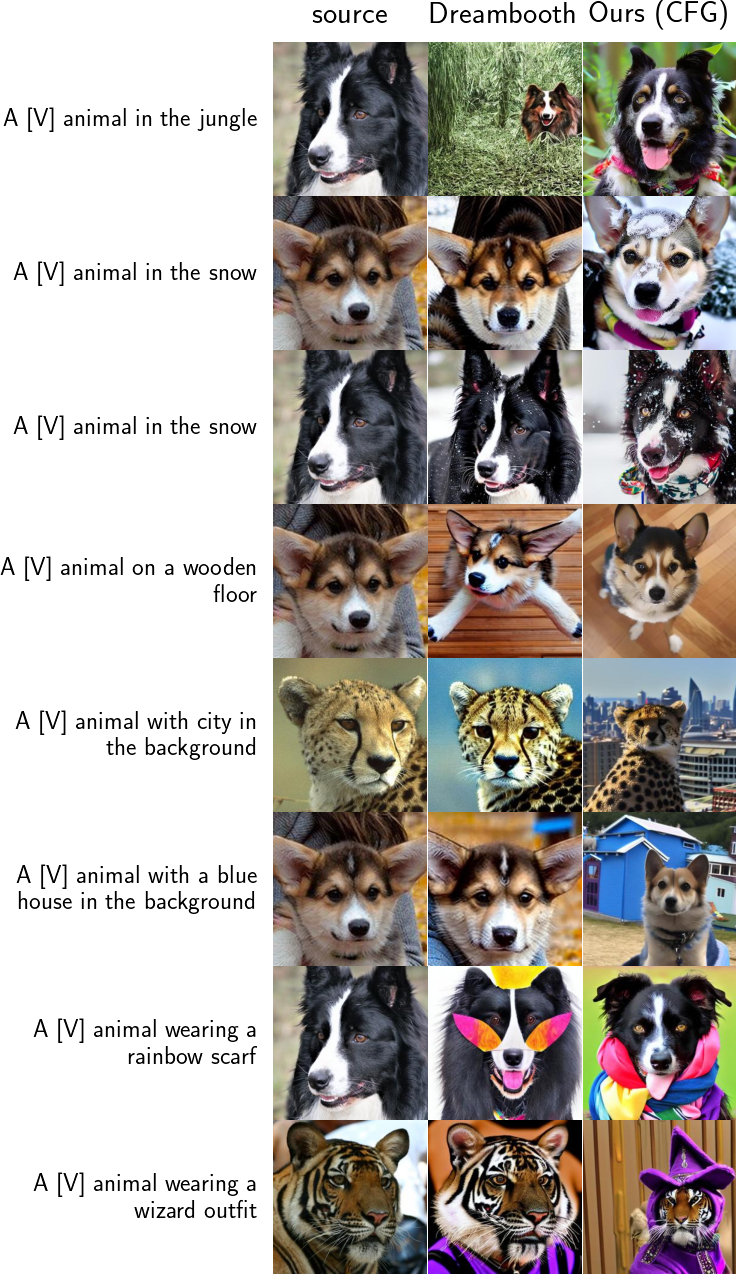}
        \caption{Qualitative results on AFHQ-v2 dataset.}
        \label{fig:afhq_qualitative_results}
    \end{figure}
    
    \begin{table}[h!]
    \centering
    \caption{\textbf{Baselines used for open-category personalization.}}
    \begin{tabular}{@{}ll@{}}
    \toprule
    \textbf{Method} & \textbf{Fine-tuning} \\
    \midrule
    \texttt{Textual Inversion}~\cite{gal2022textual} & Yes \\
    \texttt{DreamBooth}~\cite{ruiz2023dreambooth} & Yes \\
    \texttt{Custom Diffusion}~\cite{kumari2022customdiffusion} & Yes \\
    \texttt{$\lambda$-ECLIPSE}~\cite{patel2024lambda} & No \\
    \texttt{ELITE}~\cite{wei2023elite} & No \\
    \texttt{SSR-Encoder}~\cite{zhang2024ssr} & No \\
    \texttt{BLIP-Diffusion}~\cite{li2023blip} & No \\
    \texttt{Subject-Diffusion}~\cite{ma2024subject} & No \\
    \texttt{JeDi}~\cite{zeng2024jedi} & No \\
    \texttt{IP-Adapter-Plus}~\cite{ye2023ip} & No \\
    \texttt{PatchDPO}~\cite{huang2025patchdpo} & No \\
    \texttt{RF-Inversion}~\cite{rout2024semantic} & No \\
    \texttt{LatentUnfold}~\cite{kang2025flux} & No \\
    \bottomrule
    \end{tabular}
    \label{tab:app_baselines}
    \end{table}

    \subsubsection{Settings}
    Table \ref{tab:app_open_hyper} shows the hyperparameters used for experiments on open-category training and inference. Note that the IP-Adapter Plus used in our work consists of SDXL as the base diffusion model. We use the default inference SDE solver, i.e., Euler integration, for the SDXL model. 
    The hypernetwork shares the same frozen image encoder of the IP Adapter, i.e., pre-trained CLIP Vit-G/14 image encoder. For the weight decoder, we use the same architecture as the resampler network of IP Adapter Plus \cite{ipadaptergit}. We use an embedding dimension of 512, and a linear head per LoRA matrix is attached to the output of the resampler network.
    
    \begin{table}[h!]
    \centering
    \caption{\textbf{Hyperparameter settings used in our open-category  experiments.}}
    \begin{tabular}{@{}ll@{}}
    \toprule
    \multicolumn{2}{c}{\textbf{Training Settings}} \\
    \midrule
    Learning Rate (LR) & $1 \times 10^{-6}$ \\
    Optimizer & AdamW \\
    Weight decay & $1 \times 10^{-4}$ \\
    Batch size & $64$ \\
    $\lambda$ (regularization) & $100$ \\
    LR Scheduler & constant with warmup \\
    LR warmup steps & 500 \\
    Number of steps & $4000$ \\
    Number of H100 hours & $\approx$ 1 \\
    LoRA rank & 2 \\
    LoRA target modules & Q, K, V cross attention matrices \\
    \midrule
    \multicolumn{2}{c}{\textbf{Inference Settings}} \\
    \midrule
    Guidance scale $(w + 1)$ & $5.0$ \\
    Number of diffusion steps & $30$ \\
    IP Adapter Scale & 0.55 \\
    \bottomrule
    \end{tabular}
    \label{tab:app_open_hyper}
    \end{table}

    \begin{figure*}[t]
        \centering
        \includegraphics[width=0.9\linewidth]{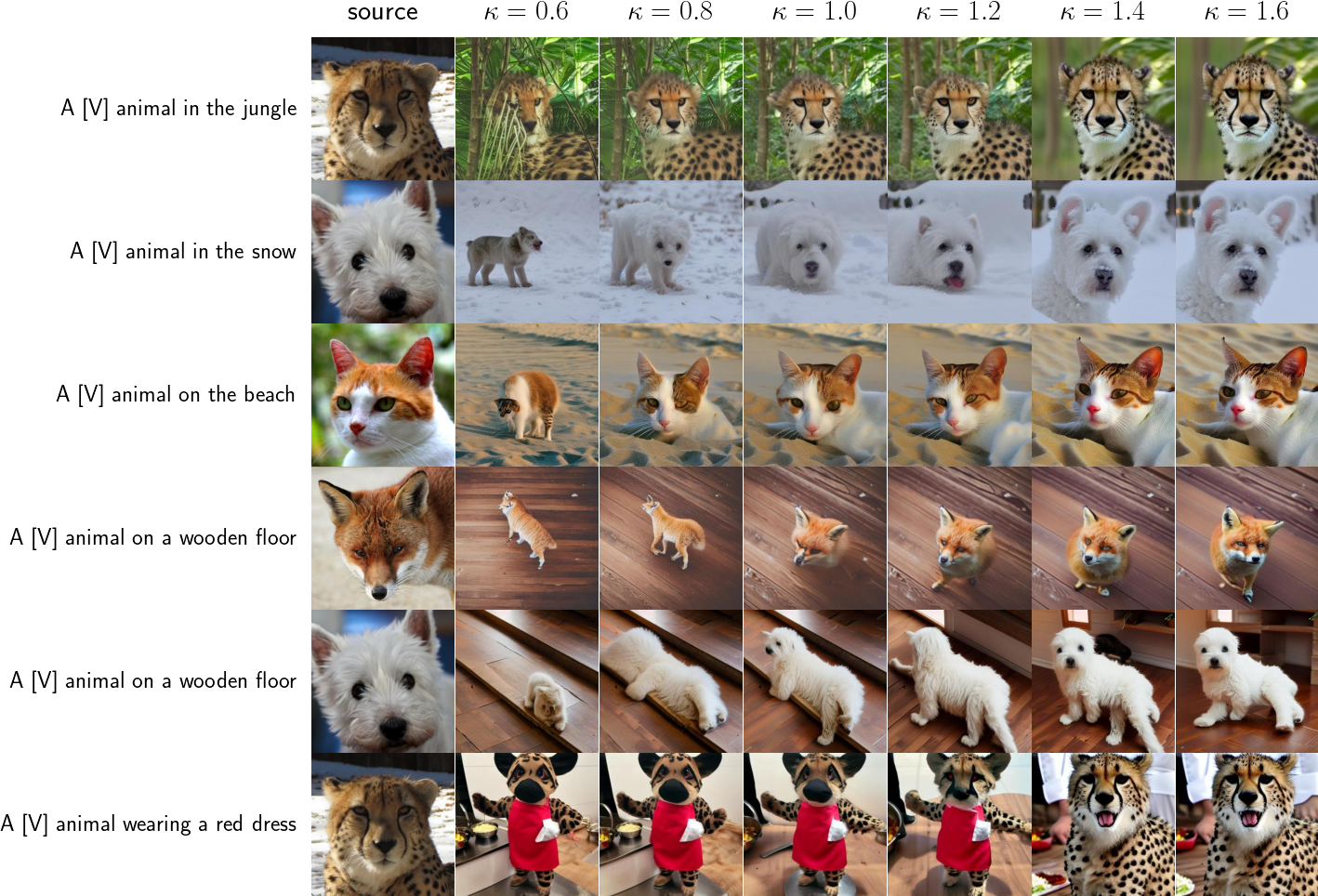}
        \caption{Qualitative results of varying $\kappa$ for the HM-CFG based sampling on AFHQ-v2 dataset.}
        \label{fig:open_control}
    \end{figure*}

    \section{Additional Results}
    
    \subsection{Qualitative results on AFHQ}
    Fig. \ref{fig:afhq_qualitative_results} shows the qualitative results of AFHQv2 compared to Dreambooth using CFG based sampling for both methods. One can see that the proposed method has better image diversity and prompt following compared to Dreambooth.

    \begin{figure*}[t]
        \centering
        \includegraphics[width=\linewidth]{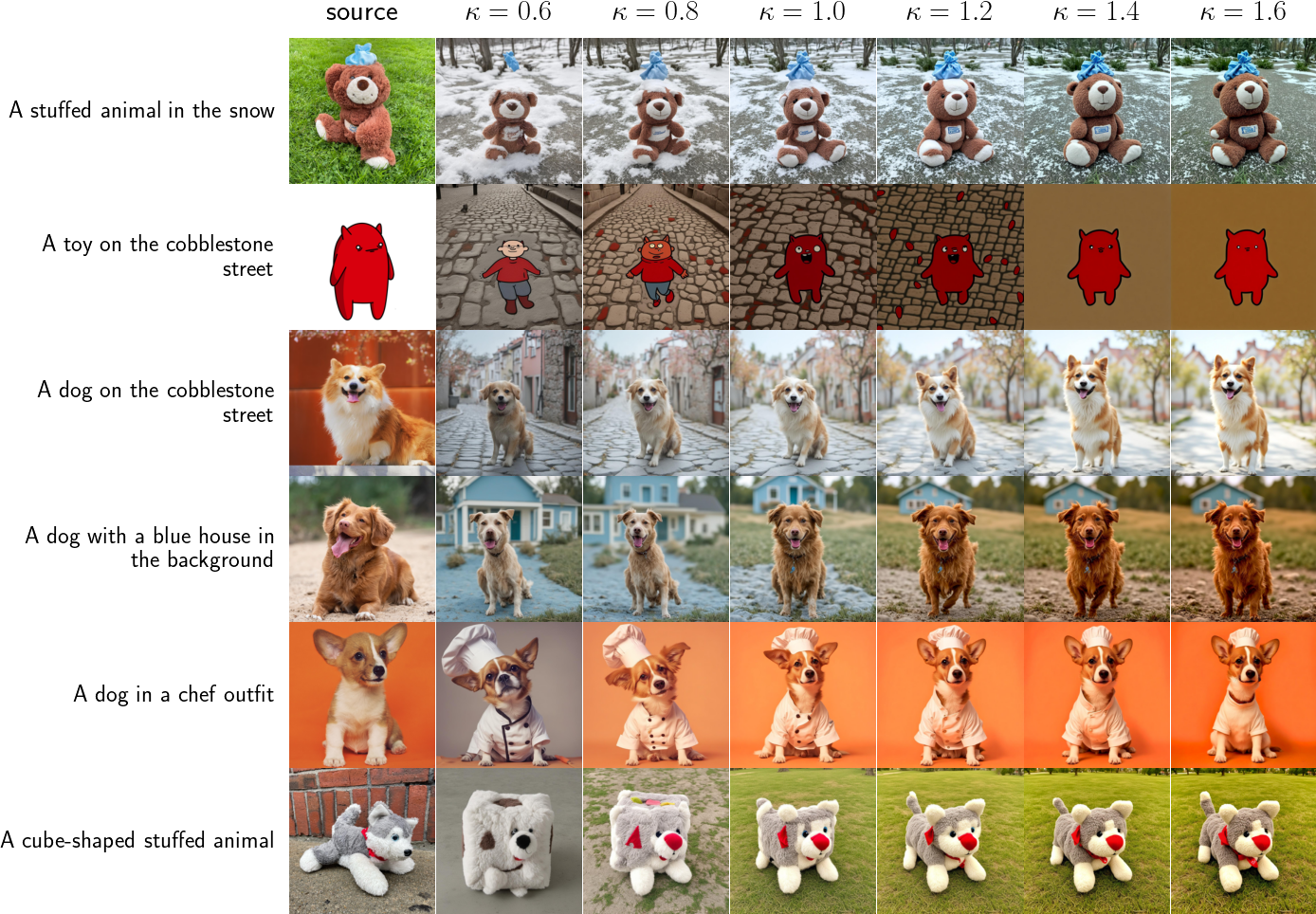}
        \caption{Qualitative results of varying $\kappa$ for the HM-CFG based sampling on Dreambench dataset.}
        \label{fig:afhq_control}
    \end{figure*}

    \subsection{Additional Results on HM-CFG}
    Table \ref{tab:app_kappa_change} shows the result of varying $\kappa$ for the AFHQv2 dataset for $w+1=7$. One can observe that changing $\kappa$ from 0 to 2 provides a tradeoff between subject and prompt fidelity. 
    Table \ref{tab:open_kappa_change} shows the result of changing $\kappa$ on the Dreambench dataset for $w+1 = 3.5$. One can see similar interpolation as in AFHQv2 dataset.

    \begin{table}[h!]
    \centering
    \caption{\textbf{Generation performance using HM-CFG approach by varying $\kappa$ on AFHQ-v2 dataset.}}
    \begin{tabular}{l|lll}
    \toprule
    $\kappa$ & CLIP-T & CLIP-I & DINO  \\
    \midrule
    CFG    & 0.278  & 0.813  & 0.717  \\
    HM-CFG ($\kappa= 0.6$)    & 0.309  & 0.732  & 0.541 \\
    HM-CFG ($\kappa= 0.8$)    & 0.304  & 0.764  & 0.631 \\
    HM-CFG ($\kappa= 1.0$)    & 0.300  & 0.783  & 0.682 \\
    HM-CFG ($\kappa= 1.2$)    & 0.296  & 0.795  & 0.711 \\
    HM-CFG ($\kappa= 1.6$)    & 0.287  & 0.811  & 0.742 \\
    \bottomrule
    \end{tabular}
    \label{tab:app_kappa_change}
    \end{table}

    \begin{table}[h!]
    \centering
    \caption{\textbf{Generation performance using HM-CFG approach by varying $\kappa$ on Dreambench dataset.}}
    \begin{tabular}{l|lll}
    \toprule
    $\kappa$ & CLIP-T & CLIP-I & DINO  \\
    \midrule
    HM-CFG ($\kappa=1.6$) & 0.297 & 0.812 & 0.696 \\
    HM-CFG ($\kappa=1.4$) & 0.303 & 0.810 & 0.693 \\
    HM-CFG ($\kappa=1.2$) & 0.310 & 0.805 & 0.689 \\
    HM-CFG ($\kappa=1.0$) & 0.317 & 0.799 & 0.679 \\
    HM-CFG ($\kappa=0.8$) & 0.325 & 0.788 & 0.660 \\
    HM-CFG ($\kappa=0.4$) & 0.342 & 0.736 & 0.543 \\
    \bottomrule
    \end{tabular}
    \label{tab:open_kappa_change}
    \end{table}

    Fig. \ref{fig:afhq_control} shows the qualitative results of varying $\kappa$ between $[0,2]$ for the AFHQv2. As expected, we can see smooth interpolation between prompt and subject fidelity by varying $\kappa$. Fig. \ref{fig:open_control} shows the qualitative results of varying $\kappa$ for the Dreambench dataset. One can see similar interpolation effect which validates our claims on inference time controllability between prompt and subject fidelity. 
    
    \end{document}


\maketitle

\appendix


\section{Details on HM-CFG}

\subsection{Diffusion Model Sampling}
Diffusion models generate samples by simulating a reverse stochastic process that transforms noise into data. This is achieved by learning to approximate the reverse of a \textit{forward diffusion process} that gradually adds noise to data.

\paragraph{Forward Process.} The forward process defines a Markov chain:
\begin{align}
    q(\x_t | \x_{t-1}) = \mathcal{N}(\x_t; \sqrt{1 - \beta_t} \x_{t-1}, \beta_t \mathbf{I}),
\end{align}
where $\beta_t$ controls the amount of Gaussian noise added at each step. After $T$ steps, $\x_T$ approximately follows a standard Gaussian distribution: $q(\x_T) \approx \mathcal{N}(\mathbf{0}, \mathbf{I})$.

\paragraph{Reverse Process and the Role of the Score Function.} To sample from the data distribution, we run the reverse process:
\begin{align}
    p_\theta(\x_{t-1} | \x_t) = \mathcal{N}(\x_{t-1}; \bm \mu_\theta(\x_t, t), \bm \Sigma_t),
\end{align}
where $\bm \mu_\theta$ is the conditional mean learned via neural networks. Under the continuous-time formulation \cite{song2020score}, the reverse process is governed by the \textit{score function}:
\begin{align}
    \s(\x_t) := \nabla_{\x_t} \log p_t(\x_t),
\end{align}
i.e., the gradient of the log-density of the noisy data at time $t$. This guides how to denoise $\x_t$ toward regions of high data density.

\paragraph{Sampling via Score-Based SDE.} A general way to sample is by solving the \textit{reverse-time stochastic differential equation} (SDE) or its deterministic counterpart (e.g., the probability flow ODE), both of which require estimating the score function $\s(\x_t)$. In practice, this is learned via training a noise predictor $\bm \epsilon_\theta(\x_t, t)$, which relates to the score as:
\begin{align}
    \s(\x_t) \approx -\frac{1}{\sigma_t} \bm \epsilon_\theta(\x_t, t),
\end{align}
where $\sigma_t$ is the standard deviation of the noise added at time $t$.

\subsubsection*{Classifier-Free Guidance (CFG).}
To condition generation on prompt $\bm c$, \textit{classifier-free guidance} (CFG) is the most commonly used sampling technique for (personalized) diffusion model, where the conditional score $\nabla_{\x_t} \log p(\x_t | \bm c)$ is boosted by interpolating with the unconditional score $\nabla_{\x_t} \log p(\x_t)$. The guided score is:
\begin{align}
    \widetilde{\s}(\x_t, \bm c) = \nabla_{\x_t} \log p(\x_t) + (w + 1) \nabla_{\x_t} \log p(\bm c | \x_t),
\end{align}
which effectively biases sampling toward images more consistent with the prompt $\bm c$.

In practice, this score is approximated using denoisers trained with and without conditioning, via:
\begin{align}
    \widetilde{\bm \epsilon}(\x_t, \bm c) = \bm \epsilon_{\bm \theta}(\x_t, \varnothing) + (w+1) \left(\bm \epsilon_{\bm \theta}(\x_t, \bm c) - \bm \epsilon_{\bm \theta}(\x_t, \varnothing) \right).
\end{align}




\subsection{Complete Derivation of HM-CFG}
Given two prompts $\bm c = \{\bm c_{\rm S}, \bm c_{\rm G}\}$, our objective is to sample from a distribution where the generated image is consistent with both the subject-specific prompt $\bm c_{\rm S}$ and the generic prompt $\bm c_{\rm G}$. Using a classifier-free guidance (CFG) approach, we modify the score used in the reverse process.

The guided score under CFG is given by:
\begin{align}
    \widetilde{\s}(\x_t, \bm c) &= \nabla_{\x_t} \log p(\x_t | \bm c) + w \nabla_{\x_t} \log p(\bm c | \x_t) \nonumber \\
    &= \nabla_{\x_t} \log p(\x_t) + (w+1)\nabla_{\x_t} \log p(\bm c | \x_t). \label{eq:cfg-score}
\end{align}

For HM-CFG, we assume conditional independence between the subject and generic prompts given $\x_t$, i.e.,
\begin{align}
    p(\bm c | \x_t) &= p(\bm c_{\rm S}, \bm c_{\rm G} | \x_t) \nonumber \\
                    &= p(\bm c_{\rm S} | \x_t) p(\bm c_{\rm G} | \x_t).
\end{align}

Now apply Bayes’ rule to each:
\begin{align}
    p(\bm c_i | \x_t) &\propto \frac{p(\x_t | \bm c_i)}{p(\x_t)}.
\end{align}
Hence, the joint becomes:
\begin{align}
    p(\bm c | \x_t) &\propto \frac{p(\x_t | \bm c_{\rm S})}{p(\x_t)} \cdot \frac{p(\x_t | \bm c_{\rm G})}{p(\x_t)} = \frac{p(\x_t | \bm c_{\rm S}) p(\x_t | \bm c_{\rm G})}{p(\x_t)^2}.
\end{align}

Taking the log and gradient:
\begin{align}
    \nabla_{\x_t} \log p(\bm c | \x_t) =& \nabla_{\x_t} \log p(\x_t | \bm c_{\rm S}) + \nabla_{\x_t} \log p(\x_t | \bm c_{\rm G})\nonumber\\
    &- 2 \nabla_{\x_t} \log p(\x_t).
\end{align}

Plugging this back into Eq.~\eqref{eq:cfg-score}:
\begin{align}
\widetilde{\s}(\x_t, \bm c) 
&= \nabla_{\x_t} \log p(\x_t) + (w+1) \left[ \nabla_{\x_t} \log p(\x_t | \bm c_{\rm S}) \right. \nonumber\\
&\quad \left. + \nabla_{\x_t} \log p(\x_t | \bm c_{\rm G}) - 2 \nabla_{\x_t} \log p(\x_t) \right]
\end{align}

Now using the approximation $\nabla_{\x_t} \log p(\x_t | \bm c) \approx -\bm \epsilon_\theta(\x_t, \bm c)/\sigma_t$ and similarly for other terms, we obtain:
\begin{align}
    \widetilde{\bm \epsilon}(\x_t, \bm c) =& \bm \epsilon_{\bm \theta_0}(\x_t, \varnothing) + (w + 1) \big( \bm \epsilon_{\bm \theta}(\x_t, \bm c_{\rm S}) + \bm \epsilon_{\bm \theta_0}(\x_t, \bm c_{\rm G}) \nonumber\\
    &- 2 \bm \epsilon_{\bm \theta_0}(\x_t, \varnothing) \big).
\end{align}

To enable tradeoff between prompt and subject fidelity, we introduce the interpolation factor $\kappa \in [0, 2]$, and generalize the formulation to:
\begin{align*}
    & \widetilde{\bm \epsilon}(\x_t, \bm c) = \bm \epsilon_{\bm \theta_0}(\x_t, \varnothing) + (w + 1) \times \\
    & \big( \kappa \bm \epsilon_{\bm \theta}(\x_t, \bm c_{\rm S}) + (2 - \kappa) \bm \epsilon_{\bm \theta_0}(\x_t, \bm c_{\rm G}) - 2 \bm \epsilon_{\bm \theta_0}(\x_t, \varnothing) \big).
\end{align*}

This completes the derivation of the Hybrid Model Classifier-Free Guidance (HM-CFG).

\section{Experimental Details}

\subsection{Closed-Category Personalization}
\subsubsection{Settings}
Table \ref{tab:app_closed_hyper} shows the list of the hyperparameter settings for the closed category personalization task for both CelebA-HQ and AFHQv2 datasets. For CelebA-HQ, we train the model for 30k steps, whereas for AFHQv2 only 4k steps appeared to be sufficient. We use the default ODE solver of SD1.5 for sampling from the personalized diffusion model. The hypernetwork architecture is the same for both CelebA-HQ and AFHQv2 experiments. Specifically, we use network architecture proposed in \cite{ruiz2023hyperdreambooth}. The image encoder is frozen ViT-B/16 \cite{dosovitskiy2020image} and the weight decoder is a 4 hidden layer transformer decoder with embedding dimension of 4. As in \cite{ruiz2023hyperdreambooth}, we use 4 self iterations of the decoder, and use a separate linear layer at the output head per each LoRA weight matrix to be predicted. 

\begin{table}[h!]
\centering
\caption{\textbf{Hyperparameter settings used in our closed category experiments.}}
\begin{tabular}{@{}ll@{}}
\toprule
\multicolumn{2}{c}{\textbf{Training Settings}} \\
\midrule
Learning Rate (LR) & $1 \times 10^{-5}$ \\
Optimizer & AdamW \\
Weight decay & $1 \times 10^{-4}$ \\
Batch size & $64$ \\
$\gamma$ & 1 \\
$\lambda$ (regularization) & $100$ \\
Scheduler & constant with warmup \\
LR warmup steps & 500 \\
Number of steps & $4000$ \\
Number of H100 hours & $\approx$ 1 \\
LoRA rank & 3 \\
LoRA target modules & Q, K, V cross attention matrices \\
\midrule
\multicolumn{2}{c}{\textbf{Inference Settings}} \\
\midrule
Guidance scale $(w + 1)$ & $7.5$ \\
Number of diffusion steps & $30$ \\
\bottomrule
\end{tabular}
\label{tab:app_closed_hyper}
\end{table}

\subsection{Open-Category Personalization}

\subsubsection{Baselines}
Table \ref{tab:app_baselines} lists all the baselines used in this work and whether they are fine-tuning based methods. The baselines were picked based on their recency, or relevance, or representativeness.

\begin{figure}[t]
    \centering
    \includegraphics[width=\linewidth]{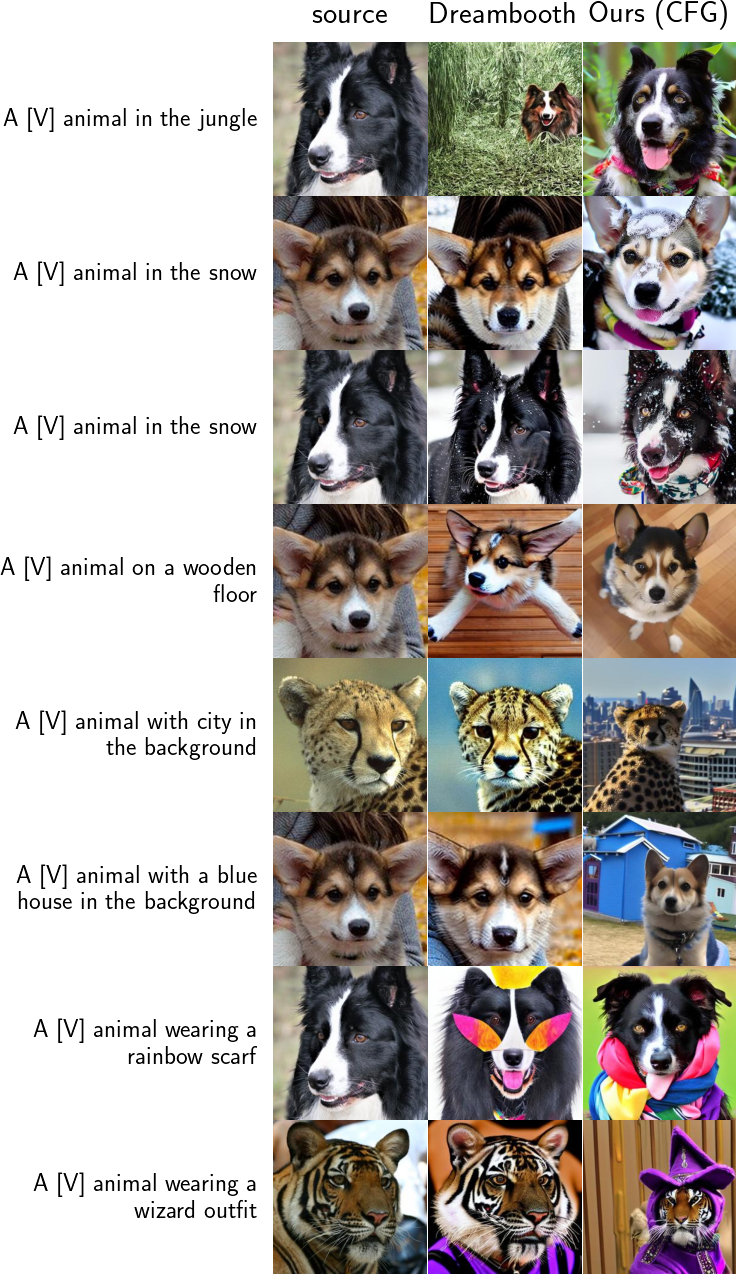}
    \caption{\textbf{Qualitative results on AFHQ-v2 dataset.}}
    \label{fig:afhq_qualitative_results}
\end{figure}

\begin{table}[h!]
\centering
\caption{\textbf{Baselines used for open-category personalization.}}
\begin{tabular}{@{}ll@{}}
\toprule
\textbf{Method} & \textbf{Fine-tuning} \\
\midrule
\texttt{Textual Inversion}~\cite{gal2022textual} & Yes \\
\texttt{DreamBooth}~\cite{ruiz2023dreambooth} & Yes \\
\texttt{Custom Diffusion}~\cite{kumari2022customdiffusion} & Yes \\
\texttt{$\lambda$-ECLIPSE}~\cite{patel2024lambda} & No \\
\texttt{ELITE}~\cite{wei2023elite} & No \\
\texttt{SSR-Encoder}~\cite{zhang2024ssr} & No \\
\texttt{BLIP-Diffusion}~\cite{li2023blip} & No \\
\texttt{Subject-Diffusion}~\cite{ma2024subject} & No \\
\texttt{JeDi}~\cite{zeng2024jedi} & No \\
\texttt{IP-Adapter-Plus}~\cite{ye2023ip} & No \\
\texttt{PatchDPO}~\cite{huang2025patchdpo} & No \\
\texttt{RF-Inversion}~\cite{rout2024semantic} & No \\
\texttt{LatentUnfold}~\cite{kang2025flux} & No \\
\bottomrule
\end{tabular}
\label{tab:app_baselines}
\end{table}

\subsubsection{Settings}
Table \ref{tab:app_open_hyper} shows the hyperparameters used for experiments on open-category training and inference. Note that the IP-Adapter Plus used in our work consists of SDXL as the base diffusion model. We use the default inference SDE solver, i.e., Euler integration, for the SDXL model. 
The hypernetwork shares the same frozen image encoder of the IP Adapter, i.e., pre-trained CLIP Vit-G/14 image encoder. For the weight decoder, we use the same architecture as the resampler network of IP Adapter Plus \cite{ipadaptergit}. We use an embedding dimension of 512, and a linear head per LoRA matrix is attached to the output of the resampler network.

\begin{table}[h!]
\centering
\caption{\textbf{Hyperparameter settings used in our open-category  experiments.}}
\begin{tabular}{@{}ll@{}}
\toprule
\multicolumn{2}{c}{\textbf{Training Settings}} \\
\midrule
Learning Rate (LR) & $1 \times 10^{-6}$ \\
Optimizer & AdamW \\
Weight decay & $1 \times 10^{-4}$ \\
Batch size & $64$ \\
$\gamma$ & 0 \\
$\lambda$ (regularization) & $100$ \\
LR Scheduler & constant with warmup \\
LR warmup steps & 500 \\
Number of steps & $4000$ \\
Number of H100 hours & $\approx$ 1 \\
LoRA rank & 2 \\
LoRA target modules & Q, K, V cross attention matrices \\
\midrule
\multicolumn{2}{c}{\textbf{Inference Settings}} \\
\midrule
Guidance scale $(w + 1)$ & $5.0$ \\
Number of diffusion steps & $30$ \\
IP Adapter Scale & 0.55 \\
\bottomrule
\end{tabular}
\label{tab:app_open_hyper}
\end{table}

\begin{figure*}[t]
    \centering
    \includegraphics[width=0.9\linewidth]{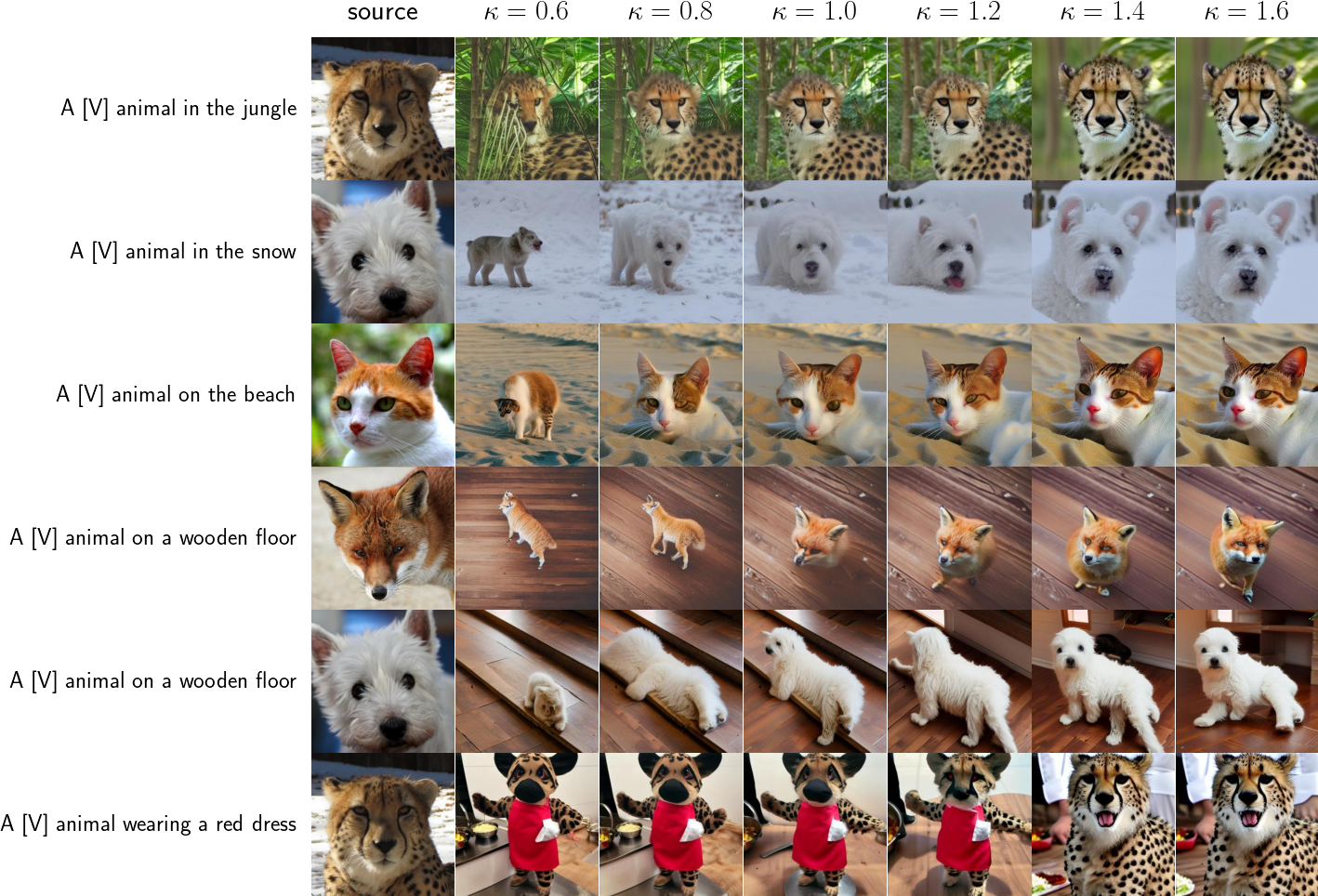}
    \caption{\textbf{Qualitative results of varying $\kappa$ for the HM-CFG based sampling on AFHQ-v2 dataset.}}
    \label{fig:open_control}
\end{figure*}

\section{Additional Results}

\subsection{Qualitative results on AFHQ}
Fig. \ref{fig:afhq_qualitative_results} shows the qualitative results of AFHQv2 compared to Dreambooth using CFG based sampling for both methods. One can see that the proposed method has better image diversity and prompt following compared to Dreambooth.

\begin{figure*}[t]
    \centering
    \includegraphics[width=\linewidth]{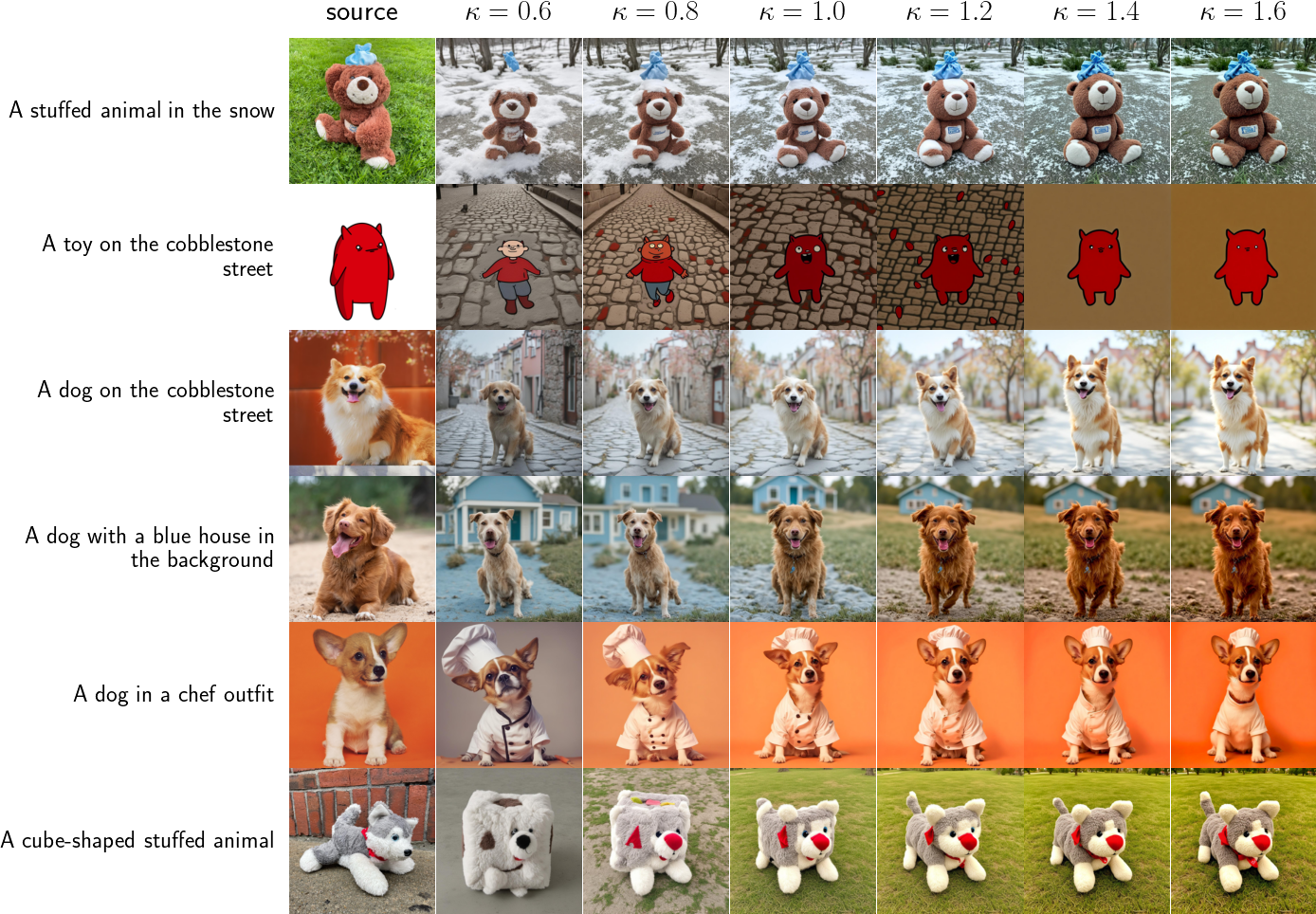}
    \caption{\textbf{Qualitative results of varying $\kappa$ for the HM-CFG based sampling on Dreambench dataset.}}
    \label{fig:afhq_control}
\end{figure*}

\subsection{Additional Results on HM-CFG}
Table \ref{tab:app_kappa_change} shows the result of varying $\kappa$ for the AFHQv2 dataset for $w+1=7$. One can observe that changing $\kappa$ from 0 to 2 provides a tradeoff between subject and prompt fidelity. 
Table \ref{tab:open_kappa_change} shows the result of changing $\kappa$ on the Dreambench dataset for $w+1 = 3.5$. One can see similar interpolation as in AFHQv2 dataset.

\begin{table}[h!]
\centering
\caption{\textbf{Generation performance using HM-CFG approach by varying $\kappa$ on AFHQ-v2 dataset.}}
\begin{tabular}{l|lll}
\toprule
$\kappa$ & CLIP-T & CLIP-I & DINO  \\
\midrule
CFG    & 0.278  & 0.813  & 0.717  \\
HM-CFG ($\kappa= 0.6$)    & 0.309  & 0.732  & 0.541 \\
HM-CFG ($\kappa= 0.8$)    & 0.304  & 0.764  & 0.631 \\
HM-CFG ($\kappa= 1.0$)    & 0.300  & 0.783  & 0.682 \\
HM-CFG ($\kappa= 1.2$)    & 0.296  & 0.795  & 0.711 \\
HM-CFG ($\kappa= 1.6$)    & 0.287  & 0.811  & 0.742 \\
\bottomrule
\end{tabular}
\label{tab:app_kappa_change}
\end{table}

\begin{table}[h!]
\centering
\caption{\textbf{Generation performance using HM-CFG approach by varying $\kappa$ on AFHQ-v2 dataset.}}
\begin{tabular}{l|lll}
\toprule
$\kappa$ & CLIP-T & CLIP-I & DINO  \\
\midrule
HM-CFG ($\kappa=1.6$) & 0.297 & 0.812 & 0.696 \\
HM-CFG ($\kappa=1.4$) & 0.303 & 0.810 & 0.693 \\
HM-CFG ($\kappa=1.2$) & 0.310 & 0.805 & 0.689 \\
HM-CFG ($\kappa=1.0$) & 0.317 & 0.799 & 0.679 \\
HM-CFG ($\kappa=0.8$) & 0.325 & 0.788 & 0.660 \\
HM-CFG ($\kappa=0.4$) & 0.342 & 0.736 & 0.543 \\
\bottomrule
\end{tabular}
\label{tab:open_kappa_change}
\end{table}

Fig. \ref{fig:afhq_control} shows the qualitative results of varying $\kappa$ between $[0,2]$ for the AFHQv2. As expected, we can see smooth interpolation between prompt and subject fidelity by varying $\kappa$. Fig. \ref{fig:open_control} shows the qualitative results of varying $\kappa$ for the Dreambench dataset. One can see similar interpolation effect which validates our claims on inference time controllability between prompt and subject fidelity. 

\bibliography{aaai2026}